\documentclass{article}

\usepackage{enumitem}
\usepackage{hyperref}
\usepackage{url}
\usepackage{comment}

\usepackage{tikz}
\usepackage{pgfplots, pgfplotstable}
\usetikzlibrary{positioning, fit, shapes.geometric, bending, decorations.text, matrix}
\usetikzlibrary{pgfplots.groupplots}
\usepgfplotslibrary{colorbrewer}
\pgfplotsset{compat=1.15}

\pgfplotsset{
    /pgfplots/legend image code/.code={%
        \draw[mark repeat=2,mark phase=2,#1] 
            plot coordinates {
                (0cm,0cm) 
                (0.19cm,0cm)
                (0.38cm,0cm)%
            };
    },
}

\usepackage{sidecap}
\usepackage{wrapfig}

\usepackage[square,numbers,sort&compress]{natbib}
\newcommand{\pgfplotLL}{
    \pgfplotsset{
        cycle list/Dark2,
        cycle multiindex* list={
            mark list*\nextlist
            Dark2\nextlist
        },
    }
}

\def\bias{{\lambda}}

\def\horizontalsep{13mm}

\def\legendscale{0.8}

\def\figwidth{0.35\linewidth}

\def\data{\mathcal{D}}

\definecolor{colorA}{HTML}{1B9E77}
\definecolor{colorB}{HTML}{D95F02}
\definecolor{colorC}{HTML}{7570B3}
\definecolor{colorD}{HTML}{E7298A}
\definecolor{colorE}{HTML}{66A61E}
\definecolor{colorF}{HTML}{E6AB02}
\definecolor{colorG}{HTML}{A6761D}
\definecolor{colorH}{HTML}{666666}

\def\colorSynthetic{colorA}
\def\colorMixed{colorB}
\def\colorFresh{colorC}

\def\synthetic{\textcolor{\colorSynthetic}{fully synthetic loop}}
\def\mixed{\textcolor{\colorMixed}{synthetic augmentation loop}}
\def\fresh{\textcolor{\colorFresh}{fresh data loop}}

\def\syntheticCap{\textcolor{\colorSynthetic}{Fully synthetic loop}}
\def\mixedCap{\textcolor{\colorMixed}{Synthetic augmentation loop}}

\usepackage[preprint]{neurips_2023}
\usepackage{LL}

\usepackage{amsmath,amsfonts,bm}

\def\eqref#1{equation~\ref{#1}}

\def\1{\bm{1}}

\def\evmu{{\mu}}

\def\mI{{\bm{I}}}

\def\mSigma{{\bm{\Sigma}}}

\DeclareMathAlphabet{\mathsfit}{\encodingdefault}{\sfdefault}{m}{sl}
\SetMathAlphabet{\mathsfit}{bold}{\encodingdefault}{\sfdefault}{bx}{n}

\definecolor{DarkGreen}{RGB}{39, 122, 43}

\def\hideNotes{0} 
\if\hideNotes0
    
    \newcommand\lolo[1]{{\color{magenta}\sf{[Lorenzo: #1]}}}
    \newcommand\josue[1]{{\color{orange}\sf{[Josue: #1]}}}
    
    \newcommand\richb[1]{{\color{blue}\sf{[richb: #1]}}}

\else
    \renewcommand{\tableofcontents}{}
    
    \newcommand\lolo[1]{}
    \newcommand\richb[1]{}
    \newcommand\josue[1]{}
\fi
\pgfplotsset{every x tick label/.append style={font=\tiny, yshift=0.5ex}}
\pgfplotsset{every y tick label/.append style={font=\tiny, xshift=0.5ex}}

\usepackage[utf8]{inputenc} 
\usepackage[T1]{fontenc}    
\usepackage{url}            
\usepackage{booktabs}       
\usepackage{amsfonts}       
\usepackage{nicefrac}       
\usepackage{microtype}      
\usepackage{xcolor}         

\hypersetup{
  citecolor=black,
}

\title{Self-Consuming Generative Models Go MAD}

\author{
{\small
Sina Alemohammad\thanks{Equal contribution.},\footnotemark[2]{~~}  Josue Casco-Rodriguez\footnotemark[1],\footnotemark[2]{~~}
 Lorenzo Luzi\footnotemark[2],~ Ahmed Imtiaz Humayun\footnotemark[2],} \\
\small \bfseries
Hossein Babaei\footnotemark[2],~ 
Daniel LeJeune\footnotemark[3],~
Ali Siahkoohi\footnotemark[4],~ Richard G.\ Baraniuk\footnotemark[2] \\
  \small {}\footnotemark[2]{~~}Department of Electrical and Computer Engineering, Rice University\\
  \small {}\footnotemark[3]{~~}Department of Statistics, Stanford University\\
  \small {}\footnotemark[4]{~~}Department of Computational Applied Mathematics and Operations Research, Rice University\\
}

\begin{document}

\maketitle

\begin{abstract}
Seismic advances in generative AI algorithms for imagery, text, and other data types has led to the temptation to use synthetic data to train next-generation models.
Repeating this process creates an 
autophagous (``self-consuming'') loop  whose properties are poorly understood.
We conduct a thorough analytical and empirical analysis using state-of-the-art generative image models of three families of autophagous loops that differ in how fixed or fresh real training data is available through the generations of training and in whether the samples from previous-generation models have been biased to trade off data quality versus diversity.  
Our primary conclusion across all scenarios is that {\em without enough fresh real data in each generation of an autophagous loop, future generative models are doomed to have their quality (precision) or diversity (recall) progressively decrease.}
We term this condition Model Autophagy Disorder (MAD), making analogy to mad cow disease.
\end{abstract}

\begin{figure}[!h]
    \centering
    \small

    \begin{tikzpicture}

        \newcommand{\imgwidth}{0.18\linewidth}


        \matrix[matrix of nodes,
                column sep=2mm,
                nodes={inner sep=0pt}]{

        Generation $t=1$ &
        $t=3$ &
        $t=5$ &
        $t=7$ &
        $t=9$
        \\ [2mm]
        \includegraphics[width=\imgwidth]{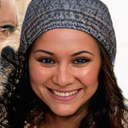} &
        \includegraphics[width=\imgwidth]{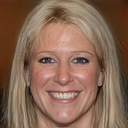} &
        \includegraphics[width=\imgwidth]{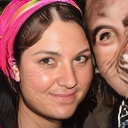} &
        \includegraphics[width=\imgwidth]{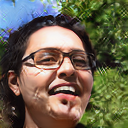} &
        \includegraphics[width=\imgwidth] {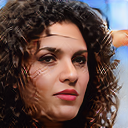}
        \\
        \includegraphics[width=\imgwidth]{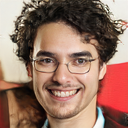} &
        \includegraphics[width=\imgwidth]{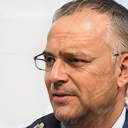} &
        \includegraphics[width=\imgwidth]{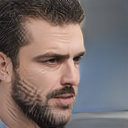} &
        \includegraphics[width=\imgwidth]{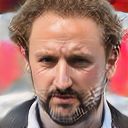} &
        \includegraphics[width=\imgwidth]
        {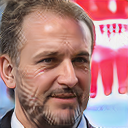}
        \\
        };

    \end{tikzpicture}
    
    \caption{
    \textbf{Training generative artificial intelligence (AI) models on synthetic data progressively amplifies artifacts.} 
As synthetic data from generative models proliferates on the Internet and in standard training datasets, future models will likely be trained on some mixture of real and synthetic data, forming an {\em autophagous (``self-consuming'') loop}.
Here we highlight one potential unintended consequence of autophagous training.
We trained a succession of StyleGAN-2~\cite{stylegan} generative models such that the training data for the model at generation $t\geq 2$ was obtained by synthesizing images from the model at generation $t-1$. 
This particular setup corresponds to a \synthetic{} in \cref{fig:loop_diagram}.
Note how the cross-hatched artifacts (possibly an architectural \textit{fingerprint}) are progressively amplified in each new generation. 
Additional samples are provided Appendices \ref{sec:ffhq_samples_unbiased} and \ref{sec:ffhq_samples_biased}.
    }
    \label{fig:ffhq_samples}
\end{figure}

\clearpage

\section{Introduction}

\subsection{Generative models are training on synthetic data from generative models}

Due to rapid advances in {\em generative artificial intelligence (AI)},  synthetic data of all kinds is rapidly proliferating. Publicly available generative models have not only revolutionized the image, audio, and text domains \cite{stable_diff, dalle2, audio_lm, music_lm, elevenai, gpt4, llama, aug_gpt}, but they are also starting to impact the creation of videos, 3D models, graphs, tables, software, and even websites \cite{dreamfusion, phenaki, digress, tabddpm, programming_is_hard, synthetic_websites}. 
Companies like Google, Microsoft, and Shutterstock are incorporating generative models into their consumer services, and the output from these services and popular generative models like Stable Diffusion \cite{stable_diff} (for images) and ChatGPT \cite{chatgpt_is_not_all_you_need} (for text) tend to end up on the Internet. The world is racing towards a future that is best summarized by a comment overheard at the 2022 ICLR conference: ``There will soon be more synthetic data than real data on the Internet.''

\begin{figure}[!t]

\centering

\begin{tikzpicture}

    \node {\includegraphics[width=0.15\linewidth]
    {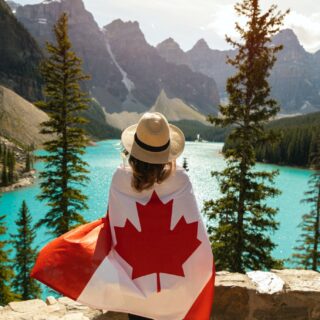}};

    \node[draw, red, line width=2mm, inner sep=0pt] at (3.6, 0)
    {\includegraphics[width=0.3\linewidth]
    {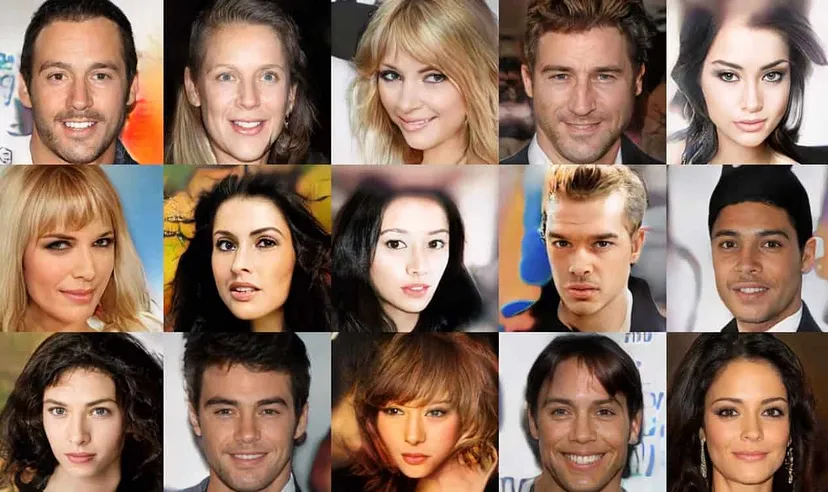}};

    \node at (7.3, 0.5) {\includegraphics[width=0.17\linewidth]
    {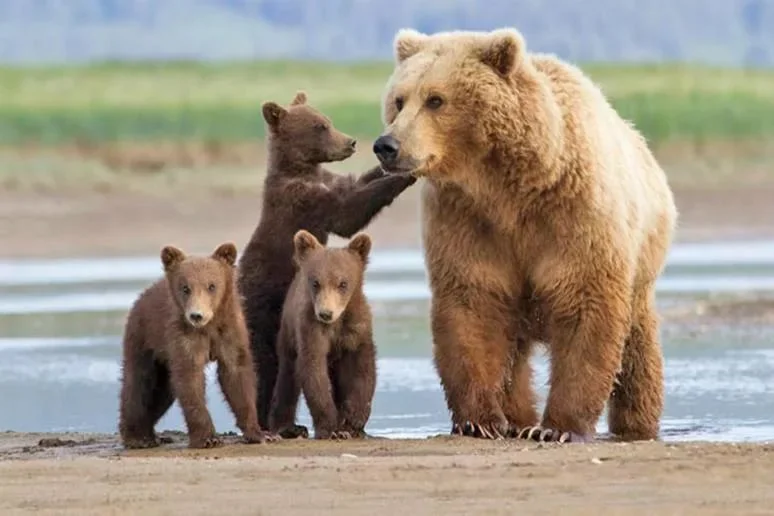}};



    \node at (10.7, -1.8)
    {\includegraphics[width=0.20\linewidth]
    {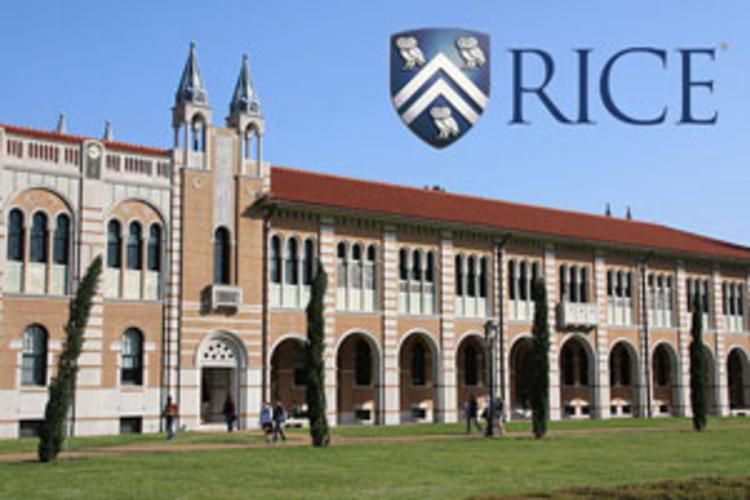}};

    \node at (0, -2) 
    {\includegraphics[width=0.13\linewidth]
    {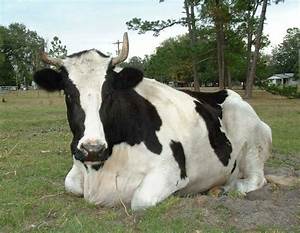}};

    \node at (2, -2.1) {\includegraphics[width=0.09\linewidth]
    {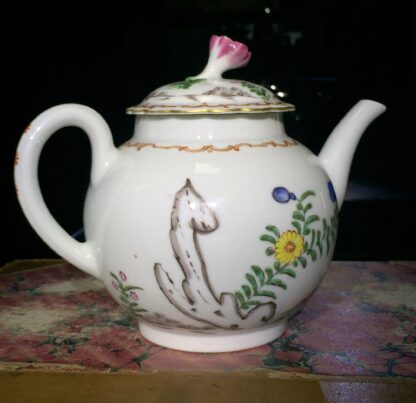}};

    \node at (3.8, -2.5) {\includegraphics[width=0.1\linewidth]
    {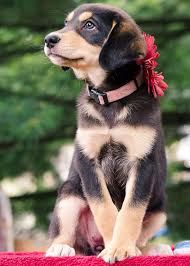}};

    \node at (5.3, -2)
    {\includegraphics[width=0.07\linewidth]
    {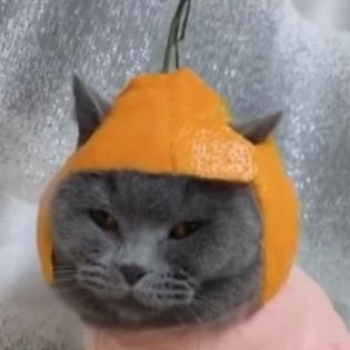}};

    \node at (7.3, -1.6) {\includegraphics[width=0.17\linewidth]
    {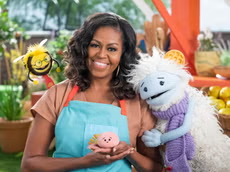}};

    \node[draw, red, line width=2mm, inner sep=0pt] at (10.6, 0.3)
    {\includegraphics[width=0.25\linewidth]
    {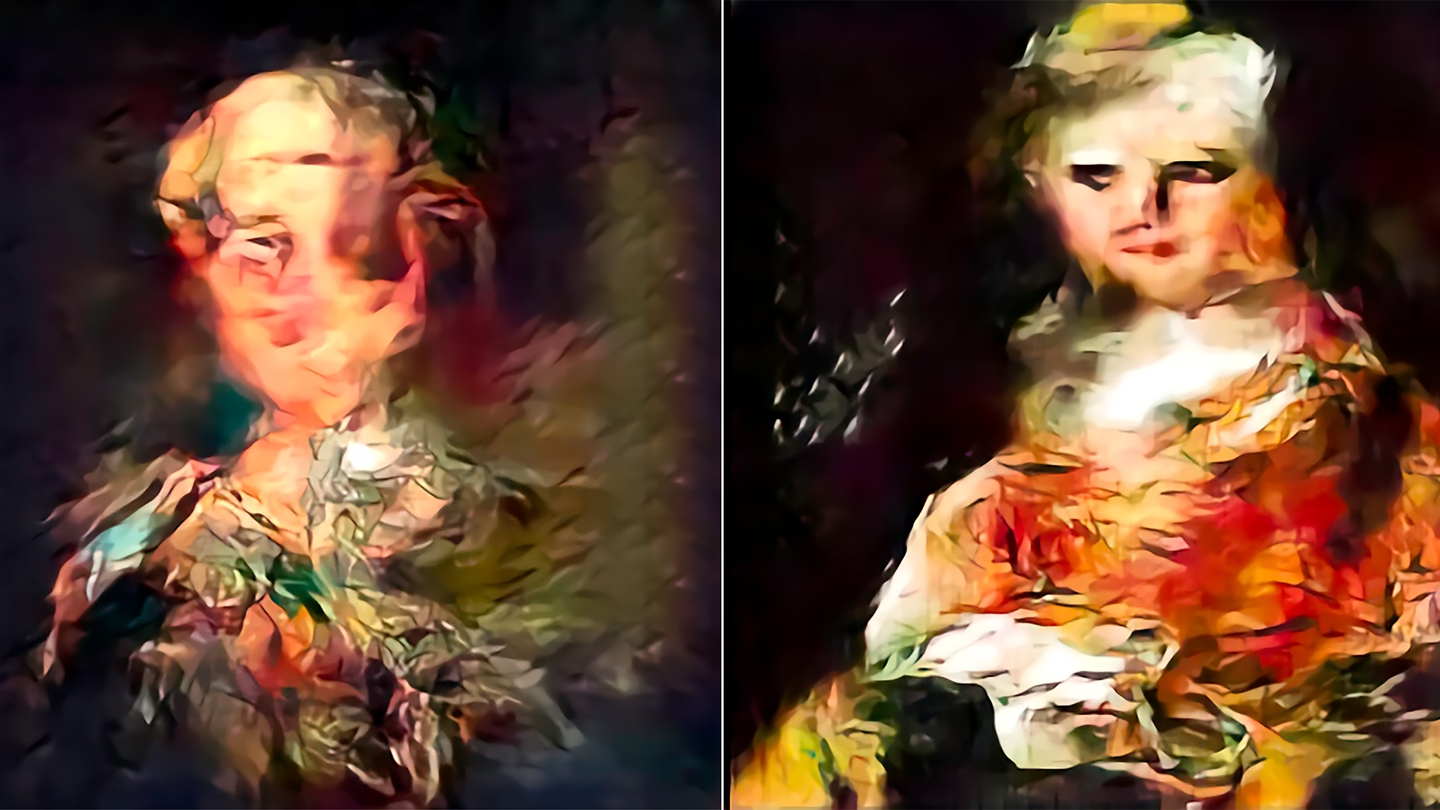}};

     \node[draw, red, line width=2mm, inner sep=0pt] at (6.7, -4)
    {\includegraphics[width=0.26\linewidth]
    {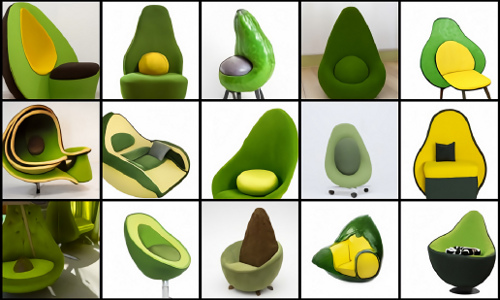}};

     \node[draw, red, line width=2mm, inner sep=0pt] at (0.7, -4.1)  {\includegraphics[width=0.3\linewidth]
    {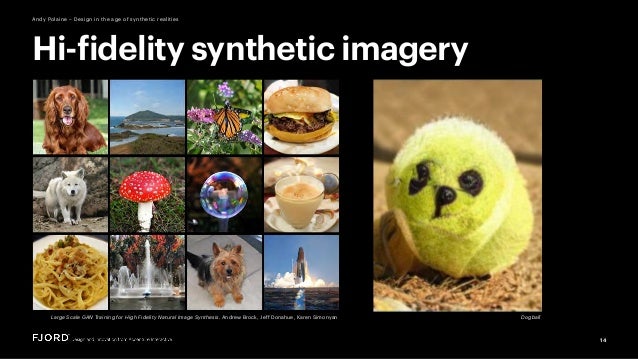}};

    \node at (3.8, -4.5)
    {\includegraphics[width=0.07\linewidth]
    {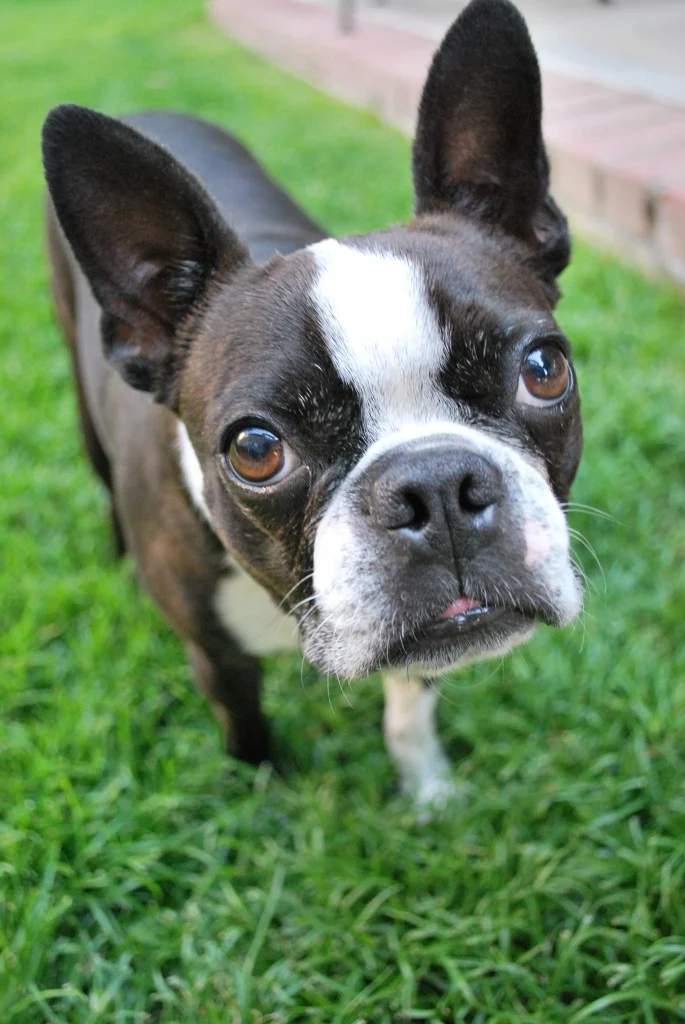}};

     \node[draw, red, line width=2mm, inner sep=0pt] at (10.7, -4)
    {\includegraphics[width=0.25\linewidth]
    {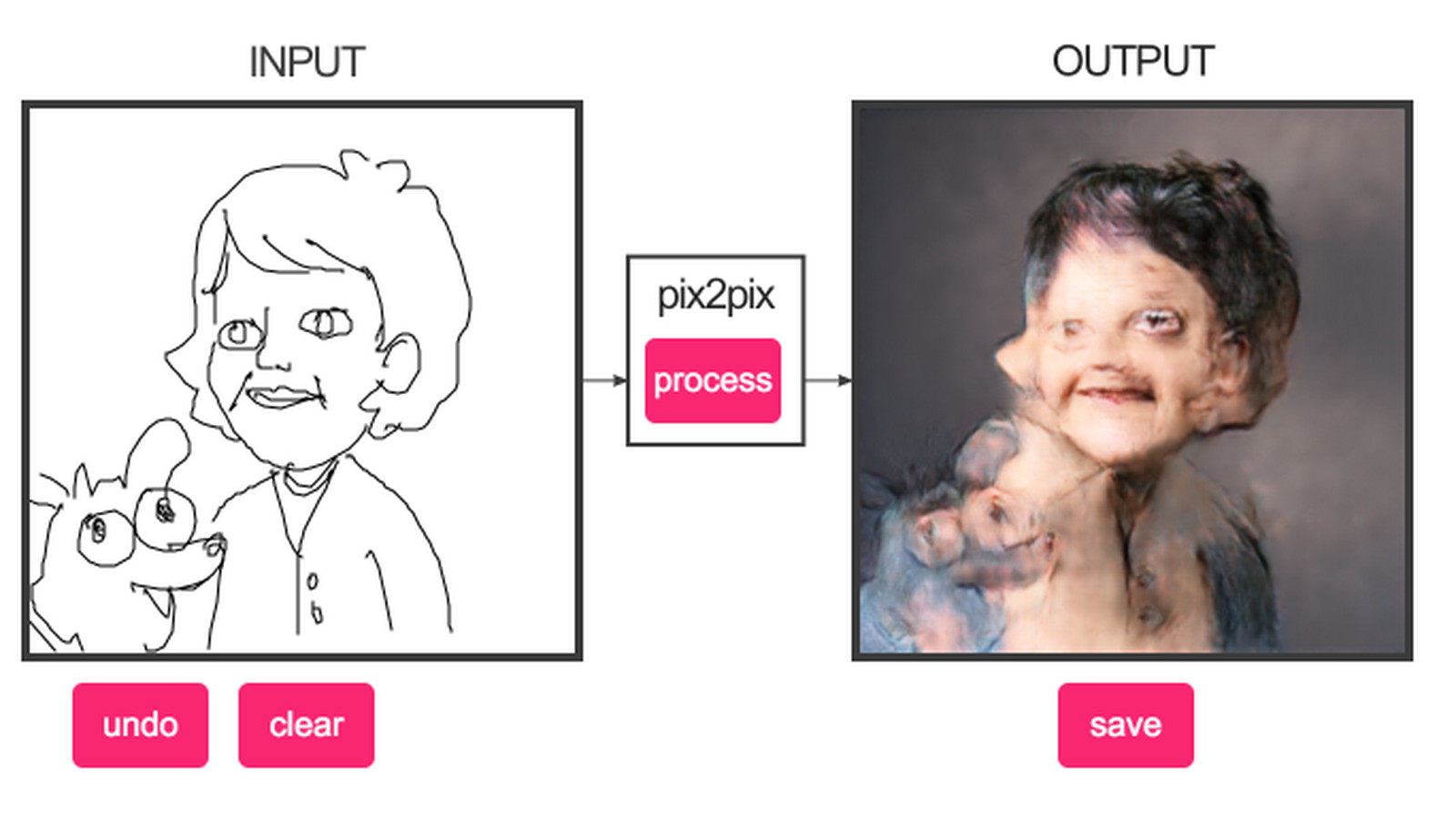}};

\end{tikzpicture}

\caption{\textbf{Today's large-scale image training datasets contain synthetic data from generative models.} Datasets such as LAION-5B \cite{laion}, which is oft-used to train text-to-image models like Stable Diffusion \cite{stable_diff}, contain synthetic images sampled from earlier generations of generative models. 
Pictured here are representative samples from LAION-5B that include (clockwise from upper left and highlighted in red) synthetic images from the generative models StyleGAN \cite{stylegan}, AICAN \cite{can}, Pix2Pix \cite{pix2pix}, DALL-E \cite{dalle}, and BigGAN \cite{biggan}. 
We found these images using simple queries on \href{https://haveibeentrained.com}{haveibeentrained.com}.
Generative models trained on the LAION-5B dataset are thus closing an autophagous (``self-consuming'') loop (see~\cref{fig:loop_diagram}) that can lead to progressively amplified artifacts (recall~\cref{fig:ffhq_samples}), 
lower quality (precision) and diversity (recall), and other unintended consequences. 
}

\label{fig:laion_imgs}
\end{figure}

Since the training datasets for generative AI models tend to be sourced from the Internet, today's AI models are unwittingly being trained on increasing amounts of AI-synthesized data.
Indeed, \cref{fig:laion_imgs} demonstrates that the popular LAION-5B dataset \cite{laion}, which is used to train state-of-the-art text-to-image models like Stable Diffusion \cite{stable_diff}, contains synthetic images sampled from several earlier generations of generative models. 
Formerly human sources of text are now increasingly created by generative AI models, from user reviews~\cite{ai_spam} to news websites~\cite{synthetic_websites}, often with no indication that the text is synthesized~\cite{cnet_secret}.
As the use of generative models continues to grow rapidly, this situation will only accelerate.

Moreover, throwing caution to the wind, AI-synthesized data is increasingly used by choice in a wide range of applications \cite{is_synthetic_data_ready, diversity_is_definitely_needed, leaving_reality_to_imagination, explore_the_power, aug_gpt,baize}, for a number of reasons.
First, it can be much easier, faster, and cheaper to synthesize training data rather than source real-world samples, particularly for data-scarce applications.
Second, in some situations synthetic data augmentation has been found empirically to boost AI system performance \cite{azizi2023synthetic, burg2023data, boomerang}. 
Third, synthetic data can protect privacy \cite{boomerang, ldfa, generation_of_anonymous_chest} in sensitive applications like medical imaging or medical record aggregation \cite{generation_of_anonymous_chest, overcoming_barriers_to_data_sharing}. 
Fourth, and most importantly, as deep learning models become increasingly enormous, we are simply running out of real data on which to train them 
\cite{bigger_is_better, large_creative_ai, will_we_run_out}. Interestingly, not only have practitioners begun deliberately training AI systems on synthetic data, but also the human annotators who provide gold-standard annotations for supervised learning tasks are increasingly using generative models to increase their productivity and income \cite{artificial_artificial_artificial}.

The witting or unwitting use of synthetic data to train generative models departs from standard AI training practice in one important respect: repeating this process for generation after generation of models forms an {\bf autophagous (``self-consuming'') loop}.
As \cref{fig:loop_diagram} details, different autophagous loop variations arise depending on how existing real and synthetic data are combined into future training sets.
Additional variations arise depending on how the synthetic data is generated. For instance, practitioners or algorithms will often introduce a {\em sampling bias} by manually ``cherry picking'' synthesized data to trade off perceptual {\em quality} (i.e., the images/texts ``look/sound good'') vs.\ {\em diversity} (i.e., many different ``types'' of images/texts are generated). 
The informal concepts of quality and diversity are closely related to the statistical metrics of {\em precision} and {\em recall}, respectively \cite{improved_precision_recall}.
If synthetic data, baised or not, is already in our training datasets today, then autophagous loops are all but inevitable in the future.

No matter what the training set makeup or sampling method, the potential ramifications of autophagous loops on the properties and performance of generative models is poorly understood.
In one direction, repeated training with synthetic data might progressively amplify the biases and artifacts present in any generative model.
We hypothesize that synthetic data contains \textit{fingerprints} of the generator architecture (e.g., convolutional traces \cite{guarnera2020deepfake} or aliasing artifacts \cite{stylegan3karras2021}) that  may be reinforced by self-consuming generators. To illustrate this, in \cref{fig:ffhq_samples} we present samples generated by StyleGAN-2 generative models after repeated training on synthetic data. 
Each generation results in a progressive amplification of cross-hatching artifacts, which are reminiscent of aliasing in StyleGAN-2 as suggested by \cite{stylegan3karras2021}. 
In another direction, autophagous loops featuring generative models tuned to produce high quality syntheses at the expense of diversity (such as \cite{stylegan, classifier_free_guidance}) might progressively dilute the diversity of the data on the Internet.
The closest exploration to this potential outcome has been the issue of {\em diversity exposure} in recommender systems,
where some studies have shown that, if a recommendation system is tuned for maximum click rate, then an echo chamber results, and users lose exposure to diverse ideas.
\cite{stroud2011niche,dylko2017dark,beam2014automating,bakshy2015exposure,o2015down}. 
Other studies have shown that, subject to the recommendation logic, the echo chamber effect might not be as pronounced \cite{brown2022echo} and could be on par with that produced by human curators \cite{moller2018not}. 
Exactly how the above and other unintended consequences could emerge from autophagous loops deserves thorough consideration and study.

For analogies and cautionary tales, one may turn to mathematics and biology.
In the language of mathematics, at one extreme, an autophagous loop is a {\em contraction mapping} that collapses to a single, boring, point, while at the other extreme it is an unstable {\em positive feedback loop} that diverges into bedlam.
Biology provides a particularly apt ``seemed like a good idea at the time'' in the practice of feeding cattle with the remains (including brains) of other cattle.
The resulting autophagous loop led to {\em mad cow disease} \cite{10.1093/oxfordjournals.aje.a009064}, a fatal neurodegenerative disease that eventually spread to humans before a massive intervention brought it under control. Lest an analogous malady disrupt the AI future, and to coin a term, it seems prudent to understand what can be done to prevent generative models from developing {\em Model Autophagy Disorder (MAD)}.

\subsection{Contributions}
\label{sec:contributions}

In this paper, we conduct a careful theoretical and empirical study of AI augophagy from the perspective of generative image models.
While we focus on image data for concreteness, the concepts developed herein apply to any data type, including text and Large Language Models (LLMs).
This paper is an elaboration of work initially published in \cite{JosueMad1,JosueMad2}; while it was being finalized, we became aware of contemporaneous work in \cite{curse_of_recursion}
and 
\cite{towards_understanding,combining_generative} that explores certain aspects of our more general theory.
We will discuss the results of these papers in context below.


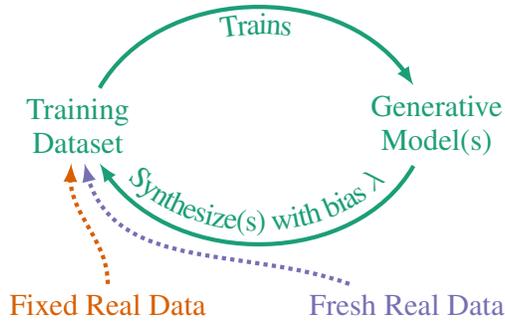
\begin{figure}[t!]
\centering

\resizebox{0.5\textwidth}{!}{
\begin{tikzpicture}

\Large

\node[align=center, text=\colorSynthetic] (a) at (-3,0) {Training\\Dataset};
\node[align=center, text=\colorSynthetic] (b) at (3,0) {Generative\\Model(s)};
\node[align=center, text=\colorMixed] (c) at (-2.5, -3) {Fixed Real Data};
\node[align=center, text=\colorFresh] (d) at (2.5, -3) {Fresh Real Data};

\draw[\colorSynthetic, line width=2pt, -latex,postaction={decorate,decoration={raise=-2ex,text along path,text
align=center,text={Trains}, text color=\colorSynthetic}}] (a) to[out=60,in=120] (b);
\draw[\colorSynthetic, line width=2pt, latex-,postaction={decorate,decoration={raise=1ex,text along path,text
align=center,text={Synthesize(s) with bias $\bias$}, text color=\colorSynthetic}}] (a) to[out=-60,in=-120] (b);

\draw[dotted, \colorMixed, line width=2pt, -latex] (c) to[out=90,in=-100] (a);
\draw[dotted, \colorFresh, line width=2pt, -latex] (d) to[out=160,in=-80] (a);


\end{tikzpicture}
}

\caption{\textbf{Recursively training generative models on synthetic data sampled from other generative models results in an autophagous (``self-consuming'') loop.}
In this paper, we study three variations of autophagous loops:
the \textit{\synthetic{}}, where a generative model is trained on only synthetic samples from previous generations (complete cycles through the loop);
the \textit{\mixed{}}, where the training set also includes a fixed set of real data;
and the \textit{\fresh{}}, where the training set also includes a fresh set of real data each generation.
See Section~\ref{sec:loops} for precise definitions.
Crucially, the generative samples are potentially obtained from a biased sampling process controlled by parameter $\bias$
that trades off sample quality vs.\ diversity.
}
\label{fig:loop_diagram}
\end{figure}

Let us summarize the three key contributions and findings that lie at the focus of this paper:

\paragraph{Realistic models for autophagous loops.} 
We propose three families of increasingly complex self-consuming training loops that realistically model the way real and synthetic data are combined into autophagous training datasets for generative models (recall \cref{fig:loop_diagram}):
\begin{itemize}
    
\item \textbf{The \synthetic{}}, wherein the training dataset for each generation's model consists solely of synthetic data sampled from previous generations' models. 
This case arises in practice, for example, when iteratively fine-tuning a generative model on its own high-quality outputs (e.g., \cite{the_power_of_synthetic_data}).
We show below in Section~\ref{sec:full} that in this case {\bf\textit{either the quality (precision) or the diversity (recall) of the generative models decreases over generations.}}
    
\item \textbf{The \mixed{}}, wherein the training dataset for each generation's model consists of a combination of synthetic data sampled from previous generations' models plus a fixed set of real training data. 
This case arises in practice, for example, in model ``self-improvement,'' where a model's training data are augmented with synthetic data from some other models (e.g., \cite{llms_self_improve}). 
We show below in Section~\ref{sec:aug} that in this case {\bf\textit{fixed real training data only delays the inevitable degradation of the quality or diversity of the generative models over generations.}}

\item \textbf{The \fresh{}}, wherein the training dataset for each generation's model consists of a combination of synthetic data sampled from previous generations' models plus a fresh set of real training data. 
This case models, for example, the standard practice where training datasets are acquired by scraping the Internet, which will find both real and synthetic data (recall \cref{fig:laion_imgs}).
We show below in Section~\ref{sec:fresh} that in this case, {\bf\textit{with enough fresh real data, the quality and diversity of the generative models do not degrade over generations.}}
    
\end{itemize}

The bottom line across all three autophagous loop models is that {\bf\em without enough fresh real data each generation, future generative models are doomed to go MAD.}

\paragraph{Sampling bias plays a key r\^{o}le in autophagous loops.} 
Users of generative models tend to manually curate (``cherry-pick'') their synthetic data, preferring high-quality samples and rejecting low-quality ones.
Moreover, state-of-the-art generative models typically feature controllable parameters that can increase synthetic quality at the expense of diversity \cite{Karras2019stylegan2, classifier_free_guidance}.
We demonstrate that the sampling biases induced through such quality-diversity (precision-recall) trade-offs have a major impact on the behavior of an autophagous training loop. 
Specifically, we show that, without sampling bias, autophagy can lead to a rapid decline of both quality and diversity, whereas, with sampling bias, quality can be maintained but diversity degrades even more rapidly.

\paragraph{Autophagous loop behaviors hold across a variety of generative models and datasets.} 
In addition to our analytical and empirical studies on simple multivariate Gaussian and Gaussian mixture models, we demonstrate in the main text and Appendix that our main conclusions hold across a variety of
generative models, including 
Denoising Diffusion Probabilistic Models (DDPM) \cite{ddpm}, 
StyleGAN-2 \cite{Karras2019stylegan2},
WGAN \cite{wgan_gp}, 
and Normalizing Flows \cite{normalizing_flow}
trained on a number of image datasets, including MNIST \cite{MNIST}
and FFHQ \cite{kass1997geometrical}.

\medskip
This paper is organized as follows.  
In Section~\ref{sec:background}, we rigorously define the concept of an autophagous loop, explain our universal biased sampling parameter $\bias$ for generative models, and define the metrics we will use to measure the quality
and diversity
of a generative model.
Then, in Sections~\ref{sec:full}, \ref{sec:aug}, and \ref{sec:fresh}, 
we study the \synthetic{}, \mixed{}, and \fresh{} models, respectively.
We conclude with a discussion in Section~\ref{sec:discussion}.
We report on the results of numerous additional experiments in various Appendices.

\section{Self-consuming generative models} 
\label{sec:background}

Modern generative models have advanced rapidly in their ability to synthesize realistic data (signals, images, videos, text, and beyond) given a finite collection of training samples from a reference (target) probability distribution $\mathcal{P}_r$. 
As generative models have proliferated, the datasets for training new models have unwittingly (see \cite{laion} and Figure~\ref{fig:laion_imgs})  or wittingly \cite{10.1007/978-3-031-18576-2_12,deng2022openfwi, llms_self_improve, self_instruct, alpaca} begun to include increasing amounts of synthetic data in addition to ``real world'' samples from $\mathcal{P}_r$
(recall Figure~\ref{fig:loop_diagram}).\footnote{
While the term ``real'' implies non-synthetic data from the ``real-world'' (e.g., a photographic image of a natural scene), in general, real data is any data drawn from the reference distribution $\mathcal{P}_r$.
}
In this section, we propose a hierarchy of increasingly realistic models for this {\em autophagy} (self-consuming) phenomenon that will enable us to draw a number of conclusions about the potential ramifications for generative modeling as synthetic training data proliferates.

\subsection{Autophagous processes}

Consider a sequence of generative models $(\model^t)_{t\in \mathbb{N}}$, where the goal is to train each model to approximate a reference probability distribution $\mathcal{P}_r$. 
At each {\em generation} $t \in \mathbb{N}$, the model $\model^t$ is trained from scratch on the dataset $\data^t = (\data_r^t, \data_s^t)$ comprised of both $n_r^t$ {\em real data samples} $\data_r^t$ drawn from $\mathcal{P}_r$ and $n_s^t$ {\em synthetic data samples} $\data_s^t$ produced by already trained generative model(s). The first-generation model $\model^1$ is trained on purely real data, i.e., $n_s^1 = 0$.

\medskip
\begin{definition*}
    An \emph{autophagous generative process} is a sequence of distributions $(\model^t)_{t\in\mathbb{N}}$ where each generative model $\model^t$ is trained on data that includes samples from previous models $(\model^\tau)_{\tau=1}^{t-1}$.
\end{definition*}

In this work, we study cases where such a process deteriorates (goes ``MAD'') over time.
Let $\mathrm{dist}(\cdot, \cdot)$ denote some distance metric on distributions.

\medskip
\begin{definition*}
    A \emph{MAD generative process} is a sequence of distributions $(\model^t)_{t\in \mathbb{N}}$ that follows a random walk such that 
    $\mathbb{E}[\mathrm{dist}(\model^t, \mathcal{P}_r)]$ increases with $t$.
    \label{defn:mad}
\end{definition*}

\begin{claim*}
    Under mild conditions, an autophagous generative process is a MAD generative process.
\end{claim*}

By studying whether a sequence of generative models $(\model^t)_{t \in \mathbb{N}}$ forms a MAD generative process, we can gain insights into the potentially detrimental effects of training generative models on synthetic data.
Two critical aspects can drive an autophagous process MAD: The balance of real and synthetic data in the training set (Section~\ref{sec:loops}) and the manner in which synthetic data is sampled from the generative models (Section~\ref{sec:sampling}).

\subsection{Variants of autophagous processes}
\label{sec:loops}

In this paper, we study three realistic autophagous mechanisms, each of which includes synthetic data and potentially real data in a feedback loop (recall \cref{fig:loop_diagram}).
We now add some additional details to the descriptions from Section~\ref{sec:contributions}:

\begin{itemize}

\item \textbf{The \synthetic{}:} 
In this scenario, each model $\model^t$ for $t \geq 2$ is trained exclusively on synthetic data sampled from models $(\model^\tau)_{\tau=1}^{t-1}$ from previous generations, i.e., $\data^t = \data_s^t$.

\item \textbf{The \mixed{}:} In this scenario, each model $\model^t$ for $t \geq 2$ is trained on a dataset $\data^t = (\data_r,\data_s^t)$ consisting of a fixed set of real data $\data_r$ sampled from $\mathcal{P}_r$ plus synthetic data $\data_s^t$ from 
models from previous generations.

\item \textbf{The \fresh{}:} In this scenario, each model $\model^t$ for $t \geq 2$ is trained on a dataset $\data^t = (\data_r^t,\data_s^t)$ consisting of a fresh set of real data $\data_r^t$ drawn independently from $\mathcal{P}_r$ plus synthetic data $\data_s^t$ from models from previous generations.

\end{itemize}

\subsection{Biased sampling in autophagous loops}
\label{sec:sampling}

While the above three autophagous loops realistically mimic real-world generative model training scenarios that involve synthetic data, it is also critical to consider how each generation's synthetic data is produced.
In particular, not all synthetic samples from a generative model will have the same level of fidelity to the training distribution, or ``quality.'' Consequently, in many applications (e.g., art generation), practitioners ``cherry-pick'' synthetic samples based on a manual evaluation of perceived quality.
It can be argued that most of the synthetic images that one can find on the Internet today are to some degree cherry-picked based on human evaluation of perceptual quality. 
Therefore, it is critical that this notion be included in the modeling and analysis of autophagous loops.

In our modeling and analysis, we implement cherry-picking via the {\em biased sampling} methods that are commonly used in generative modeling practice, such as truncation in BigGAN and StyleGAN \cite{biggan, Karras2019stylegan2}, guidance in diffusion models \cite{classifier_free_guidance}, polarity sampling \cite{Humayun_2022_CVPR}, and temperature sampling in large language models \cite{gpt4}. 
These techniques assume that the data manifold is better approximated in the higher density regions of the learned distribution. 
By biasing a generative model's synthetic samples to be drawn from parts of the learned generative model distribution $\model^t$ that are closer to its modes, these methods increase sample fidelity or quality by trading off the variety or diversity of the synthesized data \cite{Humayun_2022_CVPR}.

We employ a number of generative models in our experiments below; each has a unique controllable parameter to 
increase sample quality. We unify these parameters in the universal {\em sampling bias parameter} $\bias \in [0, 1]$, where $\bias = 1$ corresponds to unbiased sampling and $\bias = 0$ corresponds to sampling from the modes of the generative distribution $\model^t$ with zero variance. The exact interpretation of $\bias$ differs across various models, but in general synthetic sample quality will increase and diversity decrease as $\bias$ is decreased from $1$. 
Below we provide specific definitions for $\bias$ for the various generative models we consider in this paper:

\begin{itemize}

\item \textbf{Gaussian model:} 
Our theoretical analysis and simplified experiments use a multivariate Gaussian toy model. 
To implement biased sampling at generation $t$, we estimate the mean $\mu_t$ and covariance $\mSigma_t$ of the training data and then sample from the distribution $\mathcal{N}(\mu_t, \bias\mSigma_t)$. 
As $\bias$ decreases, we draw samples closer to the mean $\mu_t$.

\item \textbf{Generative adversarial network:} 
In our StyleGAN experiments, we use the truncation parameter to increase sampling quality. 
Style-based generative networks employ a secondary latent space called the style-space. 
When using truncation during inference, latent vectors in the style-space are linearly interpolated towards the mean of the style-space latent distribution. 
We denote the truncation factor by $\bias$; as $\bias$ is decreased from $1$, samples are drawn closer to the mean of the style-space distribution. 

\item \textbf{Denoising diffusion probabilistic model (DDPM):} 
For conditional diffusion models, we use classifier-free diffusion guidance \cite{classifier_free_guidance} to bias the sampling towards higher probability regions. 
We use $10\%$ conditioning dropout during training to enable classifier-free guidance. 
We define the bias parameter $\lambda$ in terms of the guidance factor $w$ from \cite{classifier_free_guidance} as $\lambda = \frac{1}{1+w}$. When $\bias = 1$, the network acts as a conventional conditional diffusion model with no guidance. 
As $\bias$ decreases, the diffusion model samples more closely to the modes of the unconditional distribution, producing higher-quality samples.

\end{itemize}

\subsection{Metrics for MADness}

Ascertaining whether an autophagous loop has gone MAD or not (recall Definition~\ref{defn:mad}) requires that we measure how far the synthesized data distribution $\model^t$ has drifted from the true data distribution $\mathcal{P}_r$ over the generations $t$.
We use the notion of the Wasserstein distance (WD) as implemented by the Fréchet Inception Distance (FID) for this purpose.
We will also find the standard concepts of precision and recall useful for making rigorous the notions of quality and diversity, respectively. 

\textbf{Wasserstein distance (WD)}, or earth mover's or optimal transport distance \cite{kantorovich1960mathematical}, measures the minimum work required to move the probability mass of one distribution to another. 
Computing the Wasserstein distance between two datasets (e.g., real and synthetic images) is prohibitively expensive. 
As such, standard practice employs the Fréchet Inception Distance (FID) \cite{fid} as an approximation, which calculates the Wasserstein-2 distance between inception feature distributions of real and synthetic images.

\textbf{Precision} quantifies the portion of synthesized samples that are deemed high {\em quality} or visually appealing. 
We use precision as an indicator of sample quality. 
We compute precision by calculating the fraction of synthetic samples that are closer to a real data example than to their $k$-th nearest neighbor \cite{improved_precision_recall}. We use $k=5$ in all experiments.

\textbf{Recall} estimates the fraction of samples in a reference distribution that are inside the support of the distribution learned by a generative model. High recall scores suggest that the generative model captures a large portion of {\em diverse} samples from the reference distribution. We compute recall in a manner similar to precision~\cite{improved_precision_recall}. 
Given a set of synthetic samples from the generative model, we calculate the fraction of real data samples that are closer to any synthetic sample than its $k$-th nearest neighbor.

\subsection{Related work}
\label{sec:related}

Contemporaneous work on feedback loops in generative modeling has explored certain aspects of our more general theory that confirm our main conclusions.

In \cite{curse_of_recursion}, the authors show that, for the \synthetic{} without sampling bias, variational autoencoders (VAE) and Gaussian mixture models result in MAD generative processes. 
They also investigate training loops resembling the \mixed{} and \fresh{}, again without sampling bias, on LLMs.
However, they take a slightly different approach from ours by fine-tuning the generative model with synthetic data instead of training from scratch. 
Their studies demonstrate that both the \mixed{} and \fresh{} can result in a decline in performance in fine-tuned LLMs over generations.

In \cite{towards_understanding}, the authors focus on the \synthetic{} with sampling bias by utilizing a diffusion model with guidance and report that it prevents a drop in image quality. 
In \cite{combining_generative}, the same authors show that a \mixed{} containing a Denoising diffusion implicit model (DDIM) \cite{song2021denoising} without sampling bias leads to poor performance over generations. The results in \cite{curse_of_recursion,towards_understanding,combining_generative} report some certain facets of a MAD generative process that align with our analytical and experimental results.

\section{The \synthetic{}: Training exclusively on synthetic data leads to MADness}
\label{sec:full}

Here we thoroughly analyze the \synthetic{}, wherein each model is trained using synthesized data from the previous generations. We focus on the the inter-generational propagation of non-idealities resulting from estimation errors and sampling biases. Specifically, we pinpoint the primary source of these non-idealities and characterize the convergence of the loop. 
The simplicity of the \synthetic{} means that it does not accurately reflect the reality of generative modeling practice.
However, one specific example of this case is when generative models are fine-tuned on their own high-quality outputs \cite{the_power_of_synthetic_data}.
Nevertheless, this loop is in a sense the worst case and so offers valuable insights that can be extrapolated to the more practical autophagous loops discussed in subsequent sections.

Our analysis and experiments below support our main conclusion for the \synthetic{}, which can be summarized as {\em either the quality (precision) or the diversity (recall) of the generative models decreases over generations.}

\subsection{Warm up: Gaussian data and martingales}

In this section, we focus the \synthetic{} and a Gaussian reference distribution and show that its martingale nature makes it a MAD generative process.

Consider a reference (real data) distribution $\mathcal{P}_r = \mathcal{N}(\mu_0, \mSigma_0)$ for some ${\evmu}_0 \in \reals^d$ and $\vect{\mSigma}_0 \in \reals^{d \times d}$, and let our generation process also be Gaussian: $\model^t = \mathcal{N}(\mu_t, \mSigma_t)$. At each time $t \in \mathbb{N}$, we sample $n_s$ vectors from $\model^{t-1}$ with sampling bias $\lambda \leq 1$; that is, we draw $X_t^1, \ldots, X_t^{n_s} \overset{\mathrm{iid}}{\sim} \mathcal{N}({\mu}_{t-1}, \bias\mSigma_{t-1})$. We then use these vectors to construct the unbiased sample mean and covariance to fit the next model $\model^t$:
\begin{align}
    \mu_t = \frac{1}{n_s} \sum_{i=1}^{n_s} X_t^i,
    \quad
    \mSigma_t = \frac{1}{n_s - 1} \sum_{i=1}^{n_s} (X_t^i - \mu_t) 
    (X_t^i - \mu_t)^\top.
    \label{eq:mu-Sigma-process}
\end{align}
In this case, we also know the distributions of these parameters. We have $\mu_t \sim \mathcal{N}(\mu_{t-1}, \frac{\bias}{n_s}\mSigma_{t-1})$ and $\mSigma_{t} \sim \mathcal{W}_d(\frac{\bias}{n_s - 1} \mSigma_{t-1},n_s-1)$, with $\mathcal{W}_d$ being the Wishart distribution. The process satisfies
\begin{align}
    \mathbb{E}[\mu_t | \mu_{t-1}] = \mu_{t-1}
    \quad\text{and}\quad
    \mathbb{E}[\mSigma_t | \mSigma_{t-1}] = \bias\mSigma_{t-1},
\end{align}
which means that $\mu_t$ and $\mSigma_t$ are {\em martingale} and {\em supermartingale} processes, respectively \cite{williams1991probability}. 
Due to the uncertainty in estimation of $\mu_t$ due to the limited sample size, $\mu_t$ is a Gaussian random walk that will tend to drift from $\mu_0$ over time, randomly biasing the distribution estimate. 
Moreover, due to being a bounded supermartingale, the covariance $\mSigma_t$ is guaranteed to converge to zero. 
The proof of the following result can be found in \Cref{sec:proof-martingale}.
\begin{proposition*}
    \label{prop:martingale}
    For the random process defined in \Cref{eq:mu-Sigma-process}, for any $\bias \leq 1$, we have $\mSigma_t \xrightarrow{\mathrm{a.s.}} 0.$
\end{proposition*}
That is, when fitting a distribution to data sampled from that distribution repeatedly, not only should we expect some modal drift because of the random walk in $\mu_t$ (reduction in \emph{quality}), but we will also inevitably experience a collapse of the variance (vanishing of \emph{diversity}).

The key idea to takeaway from this is that these effects---the random walk and the variance collapse---are solely due to the estimation error of fitting the model parameters using random data. 
Importantly, this result holds true even when there is no sampling bias (i.e., $\bias =1$). 
The magnitudes of the steps of the random walk in $\mu_t$ are determined by two main factors: the number of samples $n_s$ and the covariance $\mSigma_t$. 
Unsurprisingly, the larger the $n_s$, the smaller the steps of the random walk, since there will be less estimation error. 
This will also slow the convergence of $\mSigma_t$ to $0$. 
Meanwhile, $\mSigma_t$ can be controlled using a sampling bias factor $\lambda < 1$. The smaller the choice of $\lambda$, the more rapidly $\mSigma_t$ will converge to zero, stopping the random walk of $\mu_t$ (as illustrated in \ref{fig:nf}). 
Thus, the sampling bias factor $\lambda$ provides a trade-off to  preserve quality at the expense of diversity.

It was recently shown in related work \cite{curse_of_recursion} that the expected Wasserstein-2 metric $\mathbb{E}[\mathrm{dist}(\model^t, \mathcal{P}_r)]$, or distributional distance, is increasing for this process.
This supports our conclusion that $\model^t$ is a MAD generative process.

\subsection{Experimental setups for the \synthetic{}}
\label{sec:synthetic_setup}

Here we simulate the \synthetic{} using two widely used types of deep generative models. Recall that the \synthetic{} first requires training an initial model $\model^1$ with a fully real dataset containing $n_r^1$ samples. 
In our experiments, all subsequent models $(\model^t)_{t=2}^\infty$ are trained using only $n_s^t$ synthetic samples from the immediately preceding model $\model^{t-1}$, where each synthetic sample is produced with sampling bias $\bias$. Our primary experiments are organized as follows:

\begin{figure}[t]

\centering
\begin{tikzpicture}

\pgfplotLL

\pgfplotsset{/pgfplots/group/every plot/.append style = {
    very thick
}};

\begin{groupplot}[group style = {group size = 3 by 1, horizontal sep = \horizontalsep}, width = \figwidth]

    \nextgroupplot[
        title = {},
        xmax=9,
        ylabel={\small FID},
        xlabel={\small Generations},
        axis x line*=bottom,
        axis y line*=left,
        grid,
        legend style = { nodes={scale=\legendscale, transform shape},column sep = 10pt, legend columns = -1,  legend to name = grouplegend1, text=black, cells={align=left},}]

        \addlegendimage{empty legend}
        \addlegendentry{\syntheticCap{} without sampling bias:}

        \addplot[\colorSynthetic,very thick]
        table [
            x index=0, 
            y index=1, 
            col sep=comma] {csv/ffhq/ffhq_fid_so.csv};

        \addplot[\colorSynthetic, very thick,dotted]
        table [
            x index=0, 
            y index=1, 
            col sep=comma] {csv/mnist_diff/MNIST_full_fid.csv};

        \addlegendentryexpanded{StyleGAN2 on FFHQ $\bias = 1$};
        \addlegendentryexpanded{DDPM on MNIST $\bias = 1$};

    \nextgroupplot[
        title = {},
        xmax=9,
        ylabel={\small Precision},
        xlabel={\small Generations},
        axis x line*=bottom,
        axis y line*=left,
        grid]

        \addplot[\colorSynthetic]
        table [
            x index=0, 
            y index=1, 
            col sep=comma] {csv/ffhq/ffhq_precision_so.csv};

        \addplot[\colorSynthetic, dotted]
        table [
            x index=0, 
            y index=1, 
            col sep=comma] {csv/mnist_diff/MNIST_full_precision.csv};

    \nextgroupplot[
        title = {},
        xmax=9,
        ylabel={\small Recall},
        xlabel={\small Generations},
        axis x line*=bottom,
        axis y line*=left,
        grid]
        
        \addplot[\colorSynthetic]
        table [
            x index=0, 
            y index=1, 
            col sep=comma] {csv/ffhq/ffhq_recall_so.csv};

        \addplot[\colorSynthetic, dotted]
        table [
            x index=0, 
            y index=1, 
            col sep=comma] {csv/mnist_diff/MNIST_full_recall.csv};

\end{groupplot}
\node at ($(group c2r1) + (0,-85pt)$) {\ref{grouplegend1}}; 

\end{tikzpicture}

\caption{\textbf{Training generative models exclusively on synthetic data in a \synthetic{} without sampling bias reduces both the quality and diversity of their synthetic data decreases over generations.} 
We plot the FID (left), quality (precision, middle), and diversity (recall, right) of synthetic FFHQ and MNIST images produced in \synthetic{}.}
\label{fig:synthetic_unbiased_fid_precision_recall}

\end{figure}
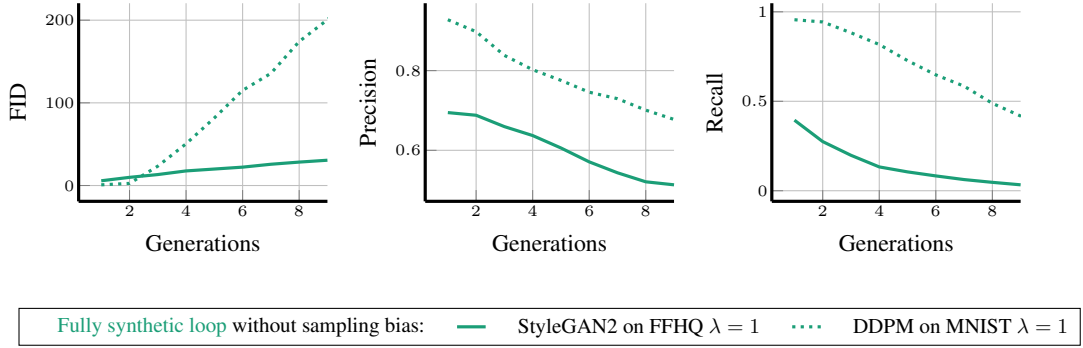

\begin{figure}[t]
    \centering 
    \foreach \b in {2,5,10,20}{
    \begin{minipage}{0.23\linewidth}
    \centering
  \small {Generation \b}   
  \includegraphics[width=\linewidth]{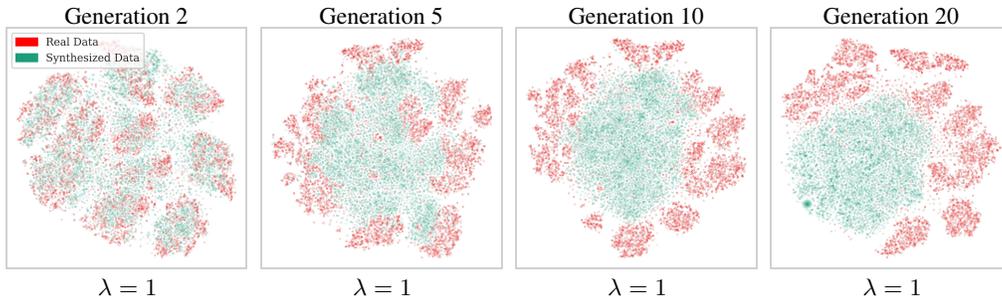} \\ \small $\bias = 1$
  \end{minipage}
  }
\caption{
\textbf{Without sampling bias, synthetic data modes drift from real modes and merge.} We present t-SNE plots of the real and synthesized data for MNIST from a \synthetic{} without sampling bias ($\bias = 1$). We can see the generated modes progressively get merged and lose separation with each other. By Generation $10$, the generated samples become almost illegible. See Figure \ref{fig:mnist_img_nobias} in the Appendix for randomly selected synthetic images of each generation.}

\label{fig:tsne1}
\end{figure}

\begin{itemize}
    \item \textbf{Denoising diffusion probabilistic model}: We use a conditional DDPM \cite{ddpm} with $T = 500$ diffusion time steps and train it on the MNIST dataset. In this experiment the synthetic dataset $\mathcal{D}_s^t$ is only sampled from the previous generation $\model^{t-1}$, with $n_r^1 = n_s^t = 60k$ for $t \ge 2$.\footnote{For all MNIST DDPM experiments we use features extracted by LeNet \cite{726791} instead of the Inception network for calculating the Wasserstein distance, since numerical digits do not fall into the domain of natural images. For consistency we also use the term ``FID'' for the MNIST results.}
    
    \item \textbf{Generative adversarial network}: We use unconditional StyleGAN2 architecture \cite{Karras2019stylegan2} and train it on the FFHQ dataset \cite{kass1997geometrical}. The images have been downsized to $128\times128$ to reduce the computational cost. Like the previous experimental setup, the synthetic samples are sampled from the previous generation with $n_r^1 = n_s^t = 70k$ for $t \ge 2$.
\end{itemize}

\begin{figure}[t]

\centering
\begin{tikzpicture}
    

\pgfplotsset{/pgfplots/group/every plot/.append style = {
    very thick
}};

    \begin{groupplot}[group style = {group size = 3 by 1, horizontal sep = \horizontalsep}, width = \figwidth]

    \nextgroupplot[
        xmax=5,
        ylabel={\small FID},
        xlabel={\small Generations},
        axis x line*=bottom,
        axis y line*=left,
        grid,
        legend style = { nodes={scale=\legendscale, transform shape},column sep = 10pt, legend columns = -1, legend to name = grouplegend2, text=black, cells={align=left},}]
        \addlegendimage{empty legend}
        \addlegendentry{\syntheticCap{} with sampling bias:}
        
        \addplot[\colorSynthetic,very thick]
        table [
            x index=0,
            y index=1,
            col sep=comma]
            {csv/ffhq/ffhq_fid_trunc.csv};
        ]

        \addplot[\colorSynthetic,very thick, dotted]
        table [
            x index=0,
            y index=4,
            col sep=comma]
            {csv/mnist_diff/MNIST_full_fid.csv};
        ]
        \addlegendentryexpanded{StyleGan on FFHQ  $\bias=0.7$};
        \addlegendentryexpanded{DDPM on MNIST  $\bias=0.5$};

    \nextgroupplot[
        xmax=5,
        ylabel={\small Precision},
        xlabel={\small Generations},
        axis x line*=bottom,
        axis y line*=left,
        grid]
        
        \addplot[\colorSynthetic]
        table [
            x index=0,
            y index=1,
            col sep=comma]
            {csv/ffhq/ffhq_precision_trunc.csv};
        ]


        \addplot[\colorSynthetic, dotted]
        table [
            x index=0,
            y index=4,
            col sep=comma]
            {csv/mnist_diff/MNIST_full_precision.csv};
        ]

    \nextgroupplot[
        xmax=5,
        ylabel={\small Recall},
        xlabel={\small Generations},
        axis x line*=bottom,
        axis y line*=left,
        grid]
        
        \addplot[\colorSynthetic]
        table [
            x index=0,
            y index=1,
            col sep=comma]
            {csv/ffhq/ffhq_recall_trunc.csv};
        ]

        \addplot[\colorSynthetic, dotted]
        table [
            x index=0,
            y index=4,
            col sep=comma]
            {csv/mnist_diff/MNIST_full_recall.csv};
        ]
        
\end{groupplot}
\node at ($(group c2r1) + (0,-85pt)$) {\ref{grouplegend2}}; 

\end{tikzpicture}

\caption{
\textbf{Training generative models on high-quality synthetic data always produces a loss in either synthetic quality or synthetic diversity. Boosting synthetic quality penalizes synthetic diversity.} We show the FID (left), quality (precision, middle), and diversity (recall, right) of synthetic FFHQ and MNIST images produced in a \synthetic{}. Values of $\bias$ less than 1 indicate that, at each iteration, synthetic diversity was traded for synthetic quality. Note that opposed to the unbiased case (\cref{fig:synthetic_unbiased_fid_precision_recall}), precision does not decay with each generation, whereas recall decays significantly faster.} 
\label{fig:synthetic_biased_fid_precision_recall}

\end{figure}
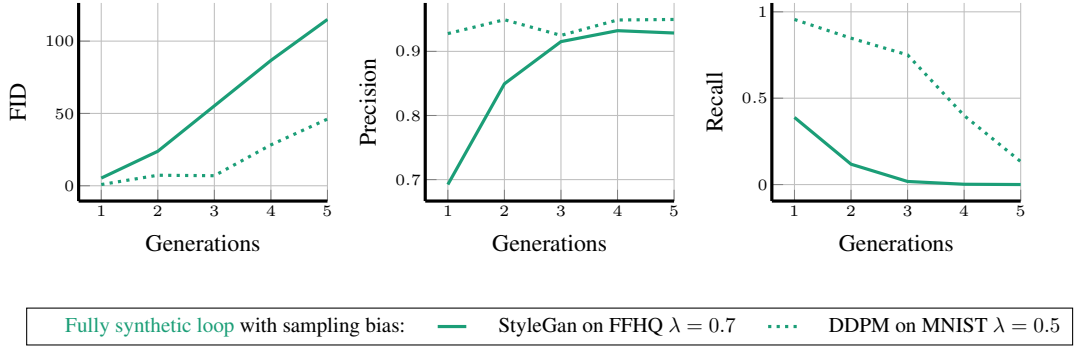

\begin{figure}[t]
    \centering 
    \foreach \b in {2,5,10,20}{
    \begin{minipage}{0.23\linewidth}
    \centering
    \small {Generation \b}
  \includegraphics[width=\linewidth]{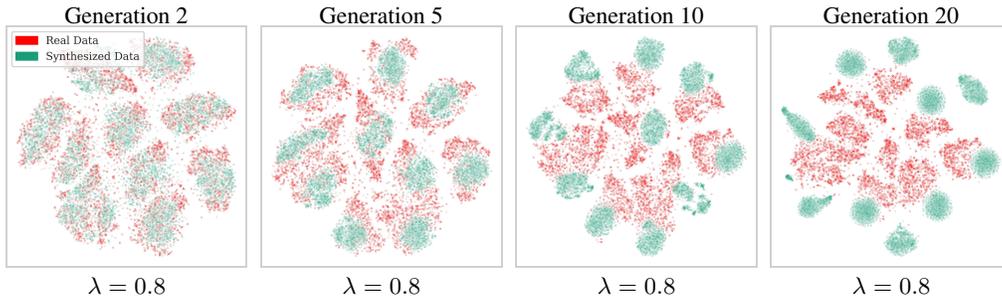} \\ \small $\bias = 0.8$
  \end{minipage}
  }
\caption{\textbf{With sampling bias, synthetic data modes drift and collapse around individual (high quality) images instead of merging.} We present t-SNE plots of the real and synthesized data for MNIST from a \synthetic{} with sampling bias ($\bias 
 = 0.8$). Note that the modes collapse onto themselves, as opposed to merging together as seen in the unbiased case (\cref{fig:tsne1}). The generated samples also remain legible. See Figure \ref{fig:mnist_img_withbias} in Appendix for randomly selected synthetic images from each generation. In \cref{sec:ffhq_samples_biased} we present qualitative examples for StyleGAN-2 where we can see that the cross-hatching artifacts do not appear but the distribution significantly loses diversity.}
\label{fig:tsne2}
\end{figure}

\subsection{Without sampling bias, the quality of synthetic data decreases}

Let us first investigate the \synthetic{} without any sampling bias ($\bias = 1$). In higher-dimensional multimodal settings, we use \textit{precision} and \textit{recall} to measure synthetic quality and diversity (as supported in \cref{sec:synthetic_gmm}). \Cref{fig:synthetic_unbiased_fid_precision_recall} illustrates the FID, precision, and recall for each generation of model. In the absence of sampling bias, the distribution of synthetic data undergoes a random walk deviating from the original distribution, caused by the finite sample size of any given training dataset. Consequently, as the generations progress, both the precision and recall of models decrease, while the FID metric exhibits a steady increase. Figure \ref{fig:mnist_gan_full_all} confirms that these trends in FID, precision, and recall continue until eventually saturating.

As the generations advance, the synthetic data distribution eventually diverges completely from the true distribution, resulting in a synthetic distribution with little resemblance to real data. This lack of realism is reflected in how the precision and recall of each model eventually drop to zero (see \Cref{fig:mnist_full_all} in the appendix for more MNIST DDPM generations), despite having a non-zero variance. 

Figure \ref{fig:tsne1} visualizes this process using the MNIST dataset. We employ the t-distributed Stochastic Neighbor Embedding (t-SNE) \cite{van2008visualizing} to reduce the dimensionality of both the real and synthetic MNIST datasets at each generation. The visualization reveals that over time, the modes of the synthetic data progressively move away from the real distribution. Despite being produced by a conditional model, these modes eventually merge together, forming a unified, larger mode of data. This gradual divergence away from the modes of real data contributes to the decrease in precision and recall, and consequently, the increase in FID, resulting in a MAD generative process.

\subsection{With biased sampling, quality can increase, but diversity will decrease rapidly}
\label{sec:synthetic_biased}

In this section, we present the results obtained with sampling bias ($\bias < 1$). Figure \ref{fig:synthetic_biased_fid_precision_recall} shows the FID, precision, and recall of models at each generations. We see that involvement of sampling bias results in increase of precision in generations; however, it causes a faster drop of recall compared to the case without sampling bias, which all together results in an increase in FID, making it a MAD generative process.

The visualization of \synthetic{}  with sampling bias is shown in Figure \ref{fig:tsne2}. In the presence of sampling bias, the movement of modes of synthetic data is confined within the support of the real data, unlike the case without sampling bias where the modes merge together. However, the variance of synthetic data rapidly decreases, resulting in very limited diversity within the synthetic data. 

We provide more experiments for the \synthetic{} with Gaussian mixture models, WGAN \cite{wgan_gp}, and Normalizing Flows \cite{normalizing_flow} in \Cref{syntheticappendix} that all result in MAD generative processes.

\begin{figure}[t]

\centering
\begin{tikzpicture}
    
\pgfplotLL

\pgfplotsset{/pgfplots/group/every plot/.append style = {
    very thick
}};

\begin{groupplot}[group style = {group size = 3 by 1, horizontal sep = \horizontalsep}, width = \figwidth]

    \nextgroupplot[
        xmax=6,
        ylabel={\small FID},
        xlabel={\small Generations},
        axis x line*=bottom,
        axis y line*=left,
        grid]

        \addplot[\colorSynthetic,very thick]
        table [
            x index=0, 
            y index=1, 
            col sep=comma] {csv/ffhq/ffhq_fid_so.csv};

        
        \addplot[\colorMixed,very thick]
        table [
            x index=0,
            y index=1,
            col sep=comma]
            {csv/ffhq/ffhq_fid_mixed.csv};
        ]

    \nextgroupplot[
        xmax=6,
        ylabel={\small Precision},
        xlabel={\small Generations},
        axis x line*=bottom,
        axis y line*=left,
        grid,
        legend style = {nodes={scale=\legendscale, transform shape}, column sep = 10pt, legend columns = -1, legend to name = grouplegend3, text=black, cells={align=left},}]

        \addlegendimage{empty legend}
        \addlegendentry{StyleGAN on FFHQ, $\bias = 1$:}

        \addplot[\colorSynthetic,very thick]
        table [
            x index=0, 
            y index=1, 
            col sep=comma] {csv/ffhq/ffhq_precision_so.csv};

        
        \addplot[\colorMixed,very thick]
        table [
            x index=0,
            y index=1,
            col sep=comma]
            {csv/ffhq/ffhq_precision_mixed.csv};
        ]
        \addlegendentryexpanded{\syntheticCap{} };
        \addlegendentryexpanded{\mixedCap{} };

    \nextgroupplot[
        xmax=6,
        ylabel={\small Recall},
        xlabel={\small Generations},
        axis x line*=bottom,
        axis y line*=left,
        grid]
        
        \addplot[\colorSynthetic]
        table [
            x index=0, 
            y index=1, 
            col sep=comma] {csv/ffhq/ffhq_recall_so.csv};

        
        \addplot[\colorMixed]
        table [
            x index=0,
            y index=1,
            col sep=comma]
            {csv/ffhq/ffhq_recall_mixed.csv};
        ]

\end{groupplot}
\node at ($(group c2r1) + (0,-85pt)$) {\ref{grouplegend3}}; 

\end{tikzpicture}

\caption{
\textbf{Training generative models in a \mixed{} with both fixed real and synthetic training data without sampling bias reduces both the quality and diversity of their synthetic data over generations, albeit more slowly than in \synthetic{} case.} We show the FID (left), quality (precision, middle), and diversity (recall, right) of synthetic FFHQ images produced in mixed-training without ($\bias= 1$) sampling bias. In \cref{sec:ffhq_samples_mixed_unbiased} we present qualitative examples, where we can see cross-hatching artifacts, similar to \cref{fig:ffhq_samples}, appearing with less prevalence.}
\label{fig:ffhq_mixed_all}

\end{figure}
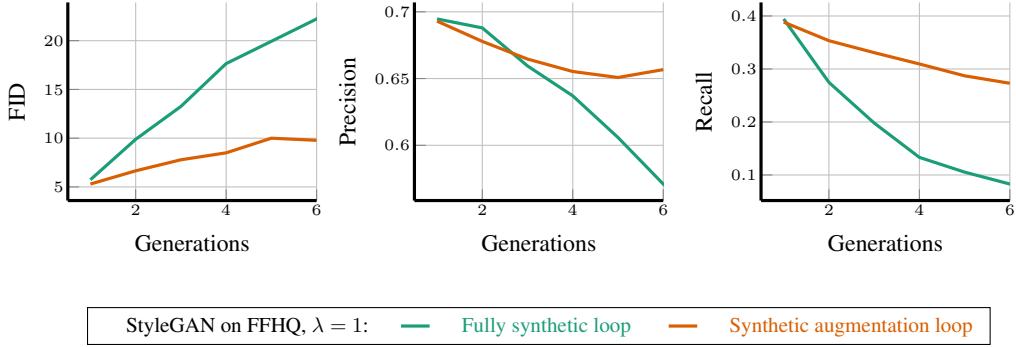

\section{The \mixed{}: Fixed real training data may delay but not prevent MADness}
\label{section4}
\label{sec:aug}

Although the analysis is tractable in the \synthetic{}, there is little reason to believe that the it will be representative of real practice. In training real generative models, practitioners will always prefer to use at least some real data when available.
In this section, we explore the \mixed{}, in which the training data consists of a fixed real dataset that is progressively augmented with synthetic data. 

We motivate the \mixed{} with the recent practice of using generative models for augmenting datasets in classification tasks, which has shown promising results
thanks to advancements in generative models
\cite{burg2023data, boomerang}. 
However, the impact of data augmentation using generative models is still not fully understood. While increasing the volume of training data generally improves the performance of machine learning models,
when synthetic samples are introduced into the dataset, there is uncertainty due to the  potential deviation of synthetic data from the true distribution of data. Even a small discrepancy can impact the model's fidelity to the real-world data distribution. As we demonstrate, the presence of the fixed real dataset is not enough to prevent this loop from producing a MAD generative process.

Our experiments below support our main conclusion for the \mixed{}, which can be summarized as {\em fixed real training data only delays the inevitable degradation of the quality or diversity of the generative models over generations.}

\subsection{Experimental setups for the \mixed{}}

Here we simulate the \mixed{} using the same deep generative models and experimental conditions as in \cref{sec:synthetic_setup}. Recall that we first require training an initial model $\model^1$ with a fully real dataset of $n_r^1$ samples. All subsequent models $(\model^t)_{t=2}^\infty$ are trained using $n_s^t$ synthetic samples from the previous model(s) and all of the original $n_r^1$ samples used to train $\model^1$. Note that each synthetic sample is always produced with sampling bias $\bias$. Our experiments are organized as follows:

\begin{itemize}
    \item \textbf{Denoising diffusion probabilistic model}: We use a conditional MNIST DDPM \cite{ddpm} with $T = 500$ diffusion time steps. In this experiment the synthetic dataset $\data_s^t$ is only sampled from the previous generation $\model^{t-1}$ with sampling bias $\bias$, and $n_r^1 = n_s^t = 60k$ for all $t \ge 2$. The original real MNIST dataset is also available at every generation: $\data_r^1 = \data_r^t$ and $n_r^1 = n_r^t = 60k$ for all $t$.
    
    \item \textbf{Generative adversarial network}: We use an unconditional StyleGAN2 architecture \cite{Karras2019stylegan2} trained on the FFHQ-128$\times$128 dataset \cite{kass1997geometrical}. Like the StyleGAN experiment in \cref{sec:synthetic_setup}, at each generation $t \ge 2$ we sample $70k$ images with no sampling bias ($\bias = 1$) from the immediately preceding model $\model^{t-1}$. However, now the synthetic dataset $\data^t_s$ includes \textit{all} the previously generated samples (not just the ones from generation $t$), producing a synthetic data pool of size $n^t_s = (t-1)70k$ that grows linearly with respect to $t$. 
    The real FFHQ dataset is always present at every generation: $\data_r^1 = \data_r^t$ and $n_r^1 = n_r^t = 70k$ for every generation $t$. 
\end{itemize}

\begin{figure}[t]

\centering
\begin{tikzpicture}
    

\pgfplotsset{/pgfplots/group/every plot/.append style = {
    very thick
}};

    \begin{groupplot}[group style = {group size = 3 by 1, horizontal sep = \horizontalsep}, width = \figwidth]

    \nextgroupplot[
        ylabel={\small FID},
        xlabel={\small Generations},
        axis x line*=bottom,
        axis y line*=left,
        grid]

        \addplot[\colorMixed]
        table [
            x index=0,
            y index=1,
            col sep=comma]
            {csv/mnist_diff/MNIST_mix_fid2.csv};
        
        \addplot[mark=o, \colorMixed]
        table [
            x index=0,
            y index=2,
            col sep=comma]
            {csv/mnist_diff/MNIST_mix_fid2.csv};
        
        \addplot[\colorMixed, densely dotted]
        table [
            x index=0,
            y index=3,
            col sep=comma]
            {csv/mnist_diff/MNIST_mix_fid2.csv};
        
        \addplot[\colorMixed, dashed]
        table [
            x index=0,
            y index=4,
            col sep=comma]
            {csv/mnist_diff/MNIST_mix_fid2.csv};

    \nextgroupplot[
        ylabel={\small Precision},
        xlabel={\small Generations},
        axis x line*=bottom,
        axis y line*=left,
        grid,
        legend style = {nodes={scale=\legendscale, transform shape}, column sep = 10pt, legend columns = -1, legend to name = grouplegend4, text=black, cells={align=left},}]

        \addlegendimage{empty legend}
        \addlegendentry{MNIST DDPM in a\\ \mixed{}:}

        \addplot[\colorMixed,very thick]
        table [
            x index=0,
            y index=1,
            col sep=comma]
            {csv/mnist_diff/MNIST_mix_precision2.csv};
        
        \addplot[mark=o, very thick, \colorMixed]
        table [
            x index=0,
            y index=2,
            col sep=comma]
            {csv/mnist_diff/MNIST_mix_precision2.csv};
        
        \addplot[\colorMixed, very thick, densely dotted]
        table [
            x index=0,
            y index=3,
            col sep=comma]
            {csv/mnist_diff/MNIST_mix_precision2.csv};
        
        \addplot[\colorMixed, very thick,dashed]
        table [
            x index=0,
            y index=4,
            col sep=comma]
            {csv/mnist_diff/MNIST_mix_precision2.csv};

        \addlegendentryexpanded{$\bias = 1$ };
        \addlegendentryexpanded{$\bias = 0.8$ };
        \addlegendentryexpanded{$\bias = 0.66$ };
        \addlegendentryexpanded{$\bias = 0.5$ };

    \nextgroupplot[
        ylabel={\small Recall},
        xlabel={\small Generations},
        axis x line*=bottom,
        axis y line*=left,
        grid]
        
        \addplot[\colorMixed]
        table [
            x index=0,
            y index=1,
            col sep=comma]
            {csv/mnist_diff/MNIST_mix_recall2.csv};
        
        \addplot[mark=o, \colorMixed]
        table [
            x index=0,
            y index=2,
            col sep=comma]
            {csv/mnist_diff/MNIST_mix_recall2.csv};
        
        \addplot[\colorMixed, densely dotted]
        table [
            x index=0,
            y index=3,
            col sep=comma]
            {csv/mnist_diff/MNIST_mix_recall2.csv};
        
        \addplot[\colorMixed, dashed]
        table [
            x index=0,
            y index=4,
            col sep=comma]
            {csv/mnist_diff/MNIST_mix_recall2.csv};


\end{groupplot}
\node at ($(group c2r1) + (0,-85pt)$) {\ref{grouplegend4}}; 

\end{tikzpicture}

\caption{\textbf{When incorporating real data in the \mixed{}, even sampling bias cannot prevent increases in FID over generations.} We show the FID (left), quality (precision, middle), and diversity (recall, right) of
synthetic MNIST images produced in a \mixed{} with different sampling biases $\bias$.
} 
\label{fig:mnist_mixed_all}

\end{figure}
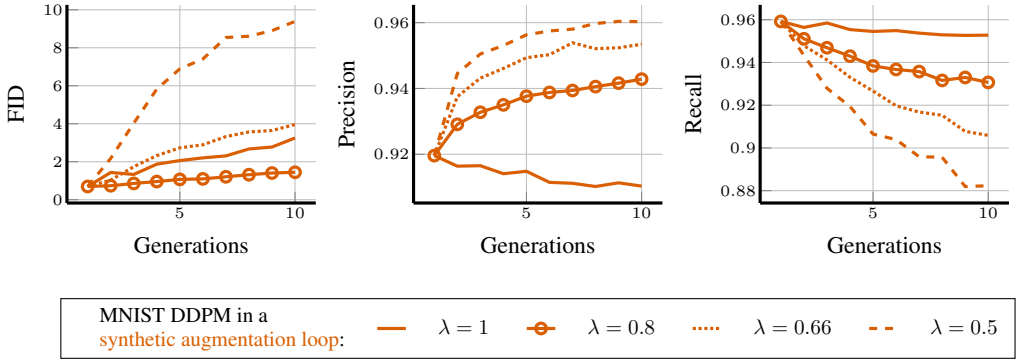

\subsection{A fixed real dataset only slows generative model degradation}

Here we show that keeping the original real dataset in the \mixed{} only slows the malignant effects of the \synthetic{} instead of preventing them. \cref{fig:ffhq_mixed_all} shows how keeping the full FFHQ dataset in a StyleGAN \mixed{} still produces the same symptoms as the \synthetic{}: the overall distance from the real dataset (FID) increases, while the quality (precision) and diversity (recall) of synthetic samples still decrease in the absence of sampling bias. In fact, in \cref{sec:ffhq_samples_mixed_unbiased} we see the same artifacts appear as in \cref{fig:ffhq_samples} and \cref{sec:ffhq_samples_unbiased}. Unlike all our other experiments, we opt for a linearly growing pool of synthetic data in the StyleGAN \mixed{} to simulate: (a) whether access to previous generations' synthesized samples could help future generations learn, and (b) what could happen to a domain of data (e.g., the Internet) in a \fresh{} with almost no newly sampled data points and unlimited access to previous generations' samples.

Additionally, \cref{fig:mnist_mixed_all} depicts how the sampling bias $\bias$ affects the \mixed{} in much the same way as it did the \synthetic{}: the overall distance from the real dataset (FID) still increases (albeit more slowly), while the synthetic quality (precision) can increase, but only at the cost of accelerated losses in synthetic diversity (recall). Naturally, some values of $\bias$ are better than others at mitigating losses in FID and precision (for example, $\bias=0.8$ in \cref{fig:mnist_mixed_all}).

\section{The \fresh{}: Fresh real data can prevent MADness}
\label{sec:fresh}

The most elaborated our autophagous loop models enable new training data to come from two sources: fresh real data from the reference distribution, and synthetic data from previously trained generative models. 
A clear instance of this can be observed in the LAION-5B dataset~\cite{laion}, which already incorporates images from generative models like Stable Diffusion~\cite{stable_diff} (recall Figure \ref{fig:laion_imgs}).

To understand the evolution of the generative models trained in this way, in this section, we investigate the \fresh{}, which takes the \mixed{} one step further by incorporating new fresh samples of real data at each iteration. 
Concretely, we imagine that the real data samples constitute only a fraction $p \in (0, 1)$ of the available data pool (e.g., a training dataset or the Internet) with the remaining fraction $1 - p$ being synthetic data from generative models. 
When we independently sample $n^t$ data points from such a training data set to train a generative model in the $t$th generation, there will be $n_r^t = p n^t$ data points that originate from the real distribution and $n_s^t = (1 - p) n^t$ synthetic data points.

In this context, we observe in our experiments below that the presence of fresh data samples fortunately mitigate the development of a MAD generative process; i.e., fresh new data helps keep the generative distribution somewhat close to the reference distribution instead of undergoing a purely random walk.
However, we still observe some alarming phenomena. 
First, we find that---regardless of the performance of early generations---the performance of later generations converges to a point that depends only on the amounts of real and synthetic data in the training loop. 
Second, we find that, while limited amounts of synthetic data can actually improve the distributional estimate in the \fresh{}---since synthetic data effectively transfers previously used real data to subsequent generations and increases the effective dataset size---too much synthetic data can still dramatically decrease the performance of the distributional estimate.

Our analysis and experiments below support our main conclusion for the \fresh{}:, which can be summarized as {\em with enough fresh real data, the quality and diversity of the generative models do not degrade over generations.}

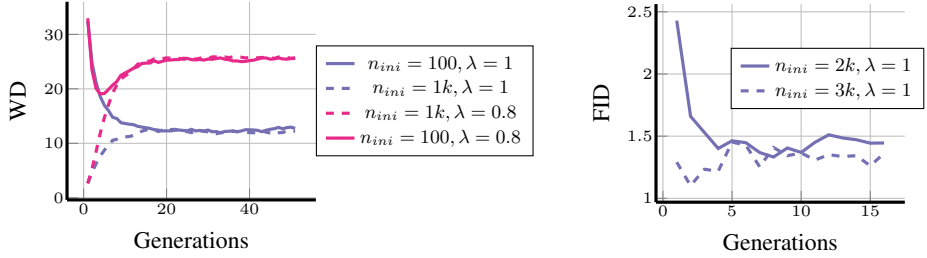
\begin{figure}[t!]
    \centering
    \begin{minipage}{\figwidth}
        \centering
        \begin{tikzpicture}

            \begin{axis}[
                ylabel={\small WD}, 
                xlabel={\small Generations},
                axis x line*=bottom,
                axis y line*=left,
                axis line style = very thick,
                grid,
                width=\linewidth,
                legend style={
                at={(1,0.5)},
                anchor=west,
                font=\footnotesize},
                legend style={nodes={scale=\legendscale, transform shape}},
                ]

                \addplot[\colorFresh, very thick]
                table [
                    x index=0, 
                    y index=1, 
                    col sep=comma] {csv/fresh/gaussian_fresh_wd.csv};
    
                \addplot[\colorFresh, very thick, dashed]
                table [
                    x index=0, 
                    y index=2, 
                    col sep=comma] {csv/fresh/gaussian_fresh_wd.csv};

                \addplot[colorD, very thick, dashed]
                table [
                    x index=0, 
                    y index=3, 
                    col sep=comma] {csv/fresh/gaussian_fresh_wd.csv};

                \addplot[colorD, very thick]
                table [
                    x index=0, 
                    y index=4, 
                    col sep=comma] {csv/fresh/gaussian_fresh_wd.csv};
    
                \addlegendentryexpanded{$n_{ini} = 100, \bias = 1$};
                \addlegendentryexpanded{$n_{ini} = 1k, \bias = 1$};
                \addlegendentryexpanded{$n_{ini} = 1k, \bias = 0.8$};
                \addlegendentryexpanded{$n_{ini} = 100, \bias = 0.8$};
                
            \end{axis}
        \end{tikzpicture}
    \end{minipage}
    \hspace{2.5cm}
    \begin{minipage}{\figwidth}
        \centering
        \begin{tikzpicture}

            \begin{axis}[
                ylabel={\small FID}, 
                xlabel={\small Generations},
                axis x line*=bottom,
                axis y line*=left,
                axis line style = very thick,
                grid,
                width=\linewidth,
                legend style={
                at={(0.3,.47)},
                anchor=south west,
                font=\footnotesize},
                legend style={nodes={scale=\legendscale, transform shape}},
                ]
    
                \addplot[\colorFresh, very thick]
                table [
                    x index=0, 
                    y index=1, 
                    col sep=comma] {csv/fresh/mnist_fresh_fid_n0.csv};
    
                \addplot[\colorFresh, very thick, dashed]
                table [
                    x index=0, 
                    y index=2, 
                    col sep=comma] {csv/fresh/mnist_fresh_fid_n0.csv};
    
                \addlegendentryexpanded{$n_{ini} = 2k, \bias = 1$};
                \addlegendentryexpanded{$n_{ini} = 3k, \bias = 1$};
                
            \end{axis}
        \end{tikzpicture}
    \end{minipage}
    \caption{\textbf{In a \fresh{}, generative models converge to a state independent of the initial generative model}. We show the Wasserstein distance (WD) and Fréchet Inception Distance (FID) of two \fresh{} models: a Gaussian model with $n_r = 100, n_s = 900$ (left) and an MNIST DDPM model with $n_r = n_s = 2k$ (right). We simulate the former with both \textcolor{\colorFresh}{unbiased} and \textcolor{colorD}{biased} sampling. Across all models we see that the asymptotic WD and FID is independent of initial real samples $n_{ini}$. }
    \label{fig:initial_point}
\end{figure}

\subsection{Experimental setups for \fresh{}}
\label{section5}

As in previous autophagous loop variants, we assume that all models are initially trained solely on real samples, with the number of real samples denoted here as $n_r^1 = n_{ini}$. In subsequent generations (i.e., for $t \geq 2$) the generative models are trained with a fixed number of real samples, denoted as $n_r^t = n_r$, and a fixed number of synthetic samples, denoted by $n_s^t = n_s$. In the \fresh{}, the dataset $\mathcal{D}_r^t$ is independently sampled from the reference probability distribution $\mathcal{P}_r$, while the dataset $\mathcal{D}_s^t$ is sampled exclusively from the previous generation $\model^{t-1}$, with a sampling bias represented as $\bias$.

Throughout the remainder of this section, we simulate the \fresh{} using different values for $n_{ini}, n_r, n_s$, and $\bias$, considering the following models and their associated reference probabilities:

\begin{itemize}
    \item \textbf{Gaussian modeling}: We consider a normal reference distribution $\mathcal{P}_r = \mathcal{N}(\bm{0}_d, \mI_d)$ with a dimension of $d = 100$. For modeling the Gaussian distribution, we utilize an unbiased moment estimation approach, as described in \Cref{eq:mu-Sigma-process}.

    \item \textbf{Denoising diffusion probabilistic model}: We use a conditional DDPM \cite{ddpm} with $T = 500$ diffusion time steps. We consider the MNIST dataset as our reference distribution. 
\end{itemize}

The Gaussian example enables examination of the \fresh{} in greater detail, especially in the asymptotic regime. Meanwhile, our MNIST DDPM example demonstrates the impact of \fresh{} on more realistic dataset and model.

\subsection{Initial models will eventually be forgotten in the \fresh{}} 

Here we investigate the impact of the initial model in the \fresh{}. We begin by training the first generative model on $n_{ini}$ samples, and train the remaining generative models with $n_r + n_s$ samples, where synthetic samples are synthesized with bias $\bias$. \Cref{fig:initial_point} summarizes the results for this experiment. 

Interestingly, for both model types, we found that the Wasserstein distance/FID converged to a limiting value after a few iterations, and that this limiting value was independent of $n_{ini}$.
In other words, for a given combination of model type and ground truth distribution $\mathcal{P}_r$, we observed that
the final outcome only depends on  $(n_r, n_s, \bias)$, that is,
\begin{align}
\lim_{t\rightarrow \infty} \mathbb{E}[\mathrm{dist}(\model^t, \mathcal{P}_r)] = \mathrm{WD}(n_r,n_s,\bias).
\label{eq:WD-limit-point}
\end{align}
Thus, the initial model's influence diminished throughout the process, with only the aforementioned parameters having an impact on the final result. 

In the context of autophagy, this point brings some hope: with the incorporation of fresh new data at each generation, there is not necessarily an increase in $\mathbb{E}[\mathrm{dist}(\model^t, \mathcal{P}_r)]$ as $t$ grows. \textit{Thus, the \fresh{} can prevent a MAD generative process.}

\subsection{A phase transition in the \fresh{}}

One might suspect that a complimentary perspective to the previous observation---that fresh new data mitigates the MAD generative process---is that synthetic data hurts a \fresh{} generative process. However, the truth appears to be more nuanced. What we find instead is that when we mix synthetic data trained on previous generations and fresh new data, there is a regime where modest amounts of synthetic data actually \emph{boost} performance, but when synthetic data exceeds some critical threshold, the models suffer. 

We make this observation precise through Gaussian simulations.
Specifically, we consider the limit point of the fresh data loop from \cref{eq:WD-limit-point}. Using the value of this limit point, which we compute via Monte-Carlo simulation, we compare against an alternative model $\model(n_e)$ trained only on a collection of real data samples of size $n_e$. We refer to $n_e$ as the \emph{effective sample size} and compute its value given $(n_r, n_s, \lambda)$ via
\begin{align}
    \mathrm{Find} \,\,\, n_e \,\,\,\, \mathrm{s.t.} \,\,\,\,\mathbb{E}[\mathrm{dist}(\model(n_e), \mathcal{P}_r)] = \mathrm{WD}(n_r,n_s,\bias).
\end{align}
That is, $n_e$ captures the sample efficiency of the limit point of the \fresh{}. We evaluate the ratio $n_e/n_r$ in our experiments. When $n_e / n_r \geq 1$, the synthetic data effectively increases the number of real samples, which we consider to be \textbf{admissible}, while for $n_e / n_r < 1$, synthetic data effectively reduces the number of real samples.

\begin{figure}[t!]
    \centering 
      \begin{minipage}{0.02\linewidth}
      \rotatebox{90}{ $\small n_s$}
    \end{minipage}
    \foreach \b in {0.7,0.85,1}{
    \begin{minipage}{0.28\linewidth}
    \centering
    $ \small \bias = \b$
    \includegraphics[width=\linewidth]{./img/fresh_data/fresh_data_lambda\b.pdf}\\
    $n_r$
  \end{minipage}
  }
  \begin{minipage}{0.056\linewidth}
    \centering
    $n_e / n_r$\\
    \includegraphics[width=\linewidth]{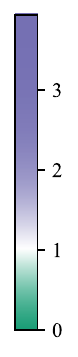}\\
    \vspace{0.6cm}

  \end{minipage}
  
\caption{\textbf{In a \fresh{}, the admissible amount of synthetic data does not increase with the amount of real data.} As the real data sample size $n_r$ increases, the maximum number of synthetic samples $n_s$ for which $n_e \geq n_r$ (\textbf{\textcolor{colorC}{blue}} area) converges. Synthetic data is only likely to be helpful when $n_r$ is small.}

\label{fig:sensitivity_lambda}
\end{figure}

\begin{figure}[th!]
    \centering 
      \begin{minipage}{0.02\linewidth}
      \rotatebox{90}{ $\small n_s$}
    \end{minipage}
    \foreach \b in {100,250,1000}{
    \begin{minipage}{0.28\linewidth}
    \centering
    $ \small n_r = \b$
    \includegraphics[width=\linewidth]{./img/fresh_data/fresh_data_nr\b.pdf}\\
    $\lambda$
  \end{minipage}
  }
  \begin{minipage}{0.0687\linewidth}
    \centering
    $n_e / n_r$\\
    \includegraphics[width=\linewidth]{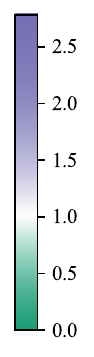}
    \\
    \vspace{0.5cm}

  \end{minipage}
  
\caption{\textbf{In a \fresh{}, sampling bias reduces the admissible synthetic sample size.} For increased sampling bias (smaller $\bias$), the maximum number of synthetic samples $n_s$ for which $n_e \geq n_r$ (\textbf{\textcolor{colorC}{blue}} area) decreases.}

\label{fig:sensitivity_nr}
\end{figure}

We plot two perspectives of the results of this experiment in \Cref{fig:sensitivity_nr,fig:sensitivity_lambda},
We discover several effects. First, we make some observations regarding sample sizes. We find that, indeed, for a given combination of $n_r$ and $\lambda < 1$, there exists a phase transition in $n_s$, such that if $n_s$ exceeds some admissible threshold, the effective sample size drops below the fresh data sample size. However, we do not find that the ratio of $n_r$ to $n_s$ is allowed to be constant; in fact, we find the opposite trend. For small values of $n_r$, we find that large value of $n_s$ can be useful, but as $n_r$ grows larger, the phase transition threshold of $n_s$ seems to become constant.

Second, we make some observations regarding the effect of sampling bias parameter $\bias$.
We find that the value of the admissible threshold for $n_s$ depends strongly on the amount of sampling bias in the synthetic process. Perhaps surprisingly, more sampling bias (smaller $\bias$) actually reduces the number of synthetic samples that can be used without harming performance. Taking the limit as $\lambda \to 1$ for unbiased sampling appears to ensure that the effective number of samples is always increased. Whether this limiting behavior extends to other generative models beyond the Gaussian modeling setting is unclear. As discussed in \Cref{sec:sampling}, it is unlikely that synthetic data is generated without sampling bias in practice, so we believe it is better to draw conclusions from the $\lambda < 1$ case.

More experiments for the \fresh{} can be found in Appendix \ref{freshappendix}.


\section{Discussion}
\label{sec:discussion}

In this paper we have sought to extrapolate what might happen in the near and distant future as generative models become ubiquitous and are used to train later generations of models in an autophagous (self-consuming) loop.
Using analysis and experiments with state-of-the-art image generative models and standard image datasets, we have studied three families of autophagous loops and singled out the key r\^{o}le played by the models' sampling bias.
Some ramifications are clear: without enough fresh real data each generation, future generative models are doomed to Model Autophagy Disorder (MAD), meaning that either their quality (measured in terms of precision) or their diversity (measured in terms of recall) will progressively degrade and generative artifacts will be amplified.
One doomsday scenario is that, if left uncontrolled for many generations, MAD could poison the data quality and diversity of the entire Internet.
Short of this, it seems inevitable that as-to-now-unseen unintended consequences will arise from AI autophagy even in the near-term.

Practitioners who are deliberately using synthetic data for training because it is cheap and easy can take our conclusions as a warning and consider tempering their synthetic data habits, perhaps by joining an appropriate 12-step program.
Those in truly data-scarce applications can interpret our results as a guide to how much scarce real data is necessary to avoid MADness in the future.
For example, future practitioners who wish to train a comprehensive medical image generator using anonymous synthetic data from multiple institutions \cite{overcoming_barriers_to_data_sharing, generation_of_anonymous_chest} should now know that very deliberate care must be taken to ensure that: (i) all anonymous synthetic images are artifact-free and diverse (see the \synthetic{}), and (ii) (ideally new) real data is present in the training as much as possible (see the \fresh{} and the \mixed{}).

Practitioners who have not been intending to use synthetic training but find it polluting their training data pool are harder to help.
To maintain trustworthy datasets containing exclusively real data, the obvious recommendation is for the community to develop methods to identify synthetic data. 
These methods can then be used to filter training datasets to reject synthetic data or maintain a particular ratio of synthetic-to-real data.
In this regard, there is early progress in the AI literature of new methods closely related to steganography \cite{guarnera2020deepfake} that can be employed for synthetic data identification.
Since generative models do not necessarily add meta-data to generated images, another approach is to {\em watermark} synthetic data so that it can be identified and rejected when training. 
The reliability of watermarking of data generated by LLMs \cite{kirchenbauer2023reliability} and novel methods for watermarking LLMs \cite{watermark}, diffusion models \cite{watermark_diffusion,peng2023protecting,wen2023tree, fernandez2023stable}, andn GANs \cite{fei2022supervised} are currently active areas of research. 
One reservation that we have about watermarking is that it deliberately introduces hidden artifacts in the synthetic data to make it detectable. 
These artifacts can possibly be amplified out of control by autophagy, turning watermarking from a useful to harmful.
In \fresh{} we see that a large amount of synthetic data hurts performance, while a modest amount of synthetic data actually boosts performance. 
Watermarking can help out in this scenario to decrease the amount of synthetic data, and ideally put the model inside the good region (e.g., the blue area in \cref{fig:sensitivity_lambda} and \cref{fig:sensitivity_nr}), such that the negative aspects of watermarking are avoided. 
This opens up interesting avenues for research on autophagy-aware watermarking.

There are many possible extensions of the work reported here, including studying combinations of the three families of autophagous loops we have proposed.
For example, one could analyze autophagous loops where the training data includes some synthetic data from previous generations' models, some fixed real data, and some fresh real data.
Our analysis has focused on the distance between the synthetic and reference data manifolds.  
An interesting research question is how this distance will manifest itself in lowered performance on AI tasks like classification
(since precision can be closely related to classifier performance, the link is waiting to be made).

Finally, in this paper we have focused on imagery, but there is nothing about our conclusions that makes them image-specific.
Generative models for any kind of data can be connected into autophagous loops and go MAD.
One timely data type is the text produced by LLMs (some of which are already being trained on synthetic data from pre-existing models like ChatGPT) \cite{llms_self_improve, self_instruct, alpaca}, where our results on precision and recall translate directly into the properties of the text produced after generations of autophagy. 
Similar conclusions have been reached in the experiments in the contemporaneous work of \cite{curse_of_recursion}, but there is much work to do in this vein.

\section*{Acknowledgements}

Thanks to Hamid Javadi, Blake Mason, and Shashank Sonkar for sharing their insights over the course of this project.
This work was supported by NSF grants CCF-1911094, IIS-1838177, and IIS-1730574; ONR grants N00014-18-12571, N00014-20-1-2534, and MURI N00014-20-1-2787; AFOSR grant FA9550-22-1-0060; DOE grant DE-SC0020345; and a Vannevar Bush Faculty Fellowship, ONR grant N00014-18-1-2047.

\clearpage

\appendix

\section{Proof of synthetic Gaussian martingale variance collapse}
\label{sec:proof-martingale}

We now prove that for the process described in \Cref{eq:mu-Sigma-process}, $\mSigma_t \xrightarrow{\mathrm{a.s.}} 0$.
\begin{proof}
First write $X_t^i = \sqrt{\lambda} \mSigma_{t-1}^{1/2} Z_t^i + \mu_{t-1}$ for $Z_t^i \sim \mathcal{N}(\bm{0}_d, \mI_d)$. 
Then consider the process $\mathrm{tr}[\mSigma_t]$, which is a lower bounded submartingale:
\begin{align}
    \mathrm{tr}[\mSigma_t] = \lambda \mathrm{tr} \left[ \mSigma_{t-1}^{1/2} \left( \frac{1}{N - 1} \sum_{i=1}^N (Z_t^i - \mu^Z_t) 
    (Z_t^i - \mu^Z_t)^\top \right) \mSigma_{t-1}^{1/2} \right],
\end{align}
where $\mu^Z_t = \frac{1}{N} \sum_{i=1}^N Z_t^i$. 
By Doob's martingale convergence theorem~\citep[Ch. 11]{williams1991probability}, there exists a random variable $W$ such that $\mathrm{tr}[\mSigma_t]\xrightarrow{\mathrm{a.s.}} W$, and we now show that we must have $W = 0$.
Without loss of generality, we can assume that $\mSigma_{t-1}$ is diagonal, in which case it becomes clear that $\mathrm{tr}[\mSigma_t]$ is a generalized $\chi^2$ random variable, being a linear combination of $d$ independent $\chi^2$ random variables with $N - 1$ degrees of freedom, mixed with weights $\lambda \mathrm{diag}(\mSigma_{t-1})$.
Therefore, we can write $\mathrm{tr}[\mSigma_t] = \lambda Y_t \mathrm{tr}[\mSigma_{t-1}]$, where $Y_t$ is a generalized $\chi^2$ random variable with the same degrees of freedom but with mixing weights $\mathrm{diag}(\mSigma_{t-1}) / \mathrm{tr}[\mSigma_{t-1}]$, and $\mathbb{E}[Y_t | \mSigma_{t-1}] = 1$. This implies that at least one mixing weight is greater than $1/D$ for each $t$, which means that for any $0 < \epsilon < 1$, there exists $c > 0$ such that $\Pr(|Y_t - 1| > \epsilon) > c$. 
Now consider the case $\lambda = 1$.
Since $|Y_t - 1| > \epsilon$ infinitely often with probability one, the only $W$ that can satisfy $\lim_{t \to \infty} \mathrm{tr}[\mSigma_0] \prod_{s=1}^t Y_s = W$ is $W = 0$. For general $\lambda \leq 1$, $\mathrm{tr}[\mSigma_t]$ is simply the product of the process for $\lambda = 1$ and the sequence $\lambda^{t - 1}$, and so the product must also converge to zero almost surely.
Finally, since $\mathrm{tr}[\mSigma_t] \xrightarrow{\mathrm{a.s.}} 0$, we also must have $\mSigma_t \xrightarrow{\mathrm{a.s.}} 0$, where convergence is defined with any matrix norm.
\end{proof}

\section{Additional experiments for the \synthetic{}}
\label{syntheticappendix}

Here we present additional experiments for the \synthetic{}.

\subsection{WGAN-GPs in an unbiased \synthetic{}}

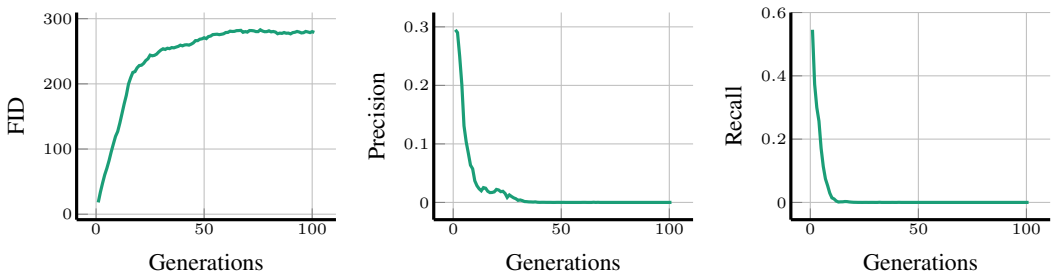
\begin{figure}[!h]

\centering
\begin{tikzpicture}
    
\pgfplotLL

\pgfplotsset{/pgfplots/group/every plot/.append style = {
    very thick, mark size=1pt, no marks
}};

\begin{groupplot}[group style = {group size = 3 by 1, horizontal sep = \horizontalsep}, width = 0.36\linewidth]

    \nextgroupplot[
        title = {},
        ylabel={\small FID},
        xlabel={\small Generations},
        axis x line*=bottom,
        axis y line*=left,
        grid]

        \addplot
        table [
            x index=0, 
            y index=1, 
            col sep=comma] {csv/mnist_gan_fid.csv};

    \nextgroupplot[
        title = {},
        ylabel={\small Precision},
        xlabel={\small Generations},
        axis x line*=bottom,
        axis y line*=left,
        grid]

        \addplot
        table [
            x index=0, 
            y index=1, 
            col sep=comma] {csv/mnist_gan_precision.csv};

    \nextgroupplot[
        title = {},
        ylabel={\small Recall},
        xlabel={\small Generations},
        axis x line*=bottom,
        axis y line*=left,
        grid]
        
        \addplot
        table [
            x index=0, 
            y index=1, 
            col sep=comma] {csv/mnist_gan_recall.csv};

\end{groupplot} 

\end{tikzpicture}

\caption{The FID (left), quality (precision, middle), and diversity (recall, right) of synthetic FFHQ and MNIST images produced by WGAN-GPs on MNIST.}
\label{fig:mnist_gan_full_all}

\end{figure}

In this experiment we trained Wasserstein GANs (with gradient penalty) \cite{wgan_gp} on the MNIST dataset in a \synthetic{} for 100 generations. As shown in \cref{fig:mnist_gan_full_all}, the FID monotonically increases, while quality (precision) and diversity (recall) monotonically decrease.

\subsection{GMMs in an unbiased \synthetic{}}
\label{sec:synthetic_gmm}

\begin{figure}[!h]
\centering

\begin{tikzpicture}

    \tikzstyle{nf_plot} = [draw, gray, line width=1mm, inner sep=0pt]

    \node[nf_plot] (original) {
    \includegraphics[width=0.2\linewidth]
    {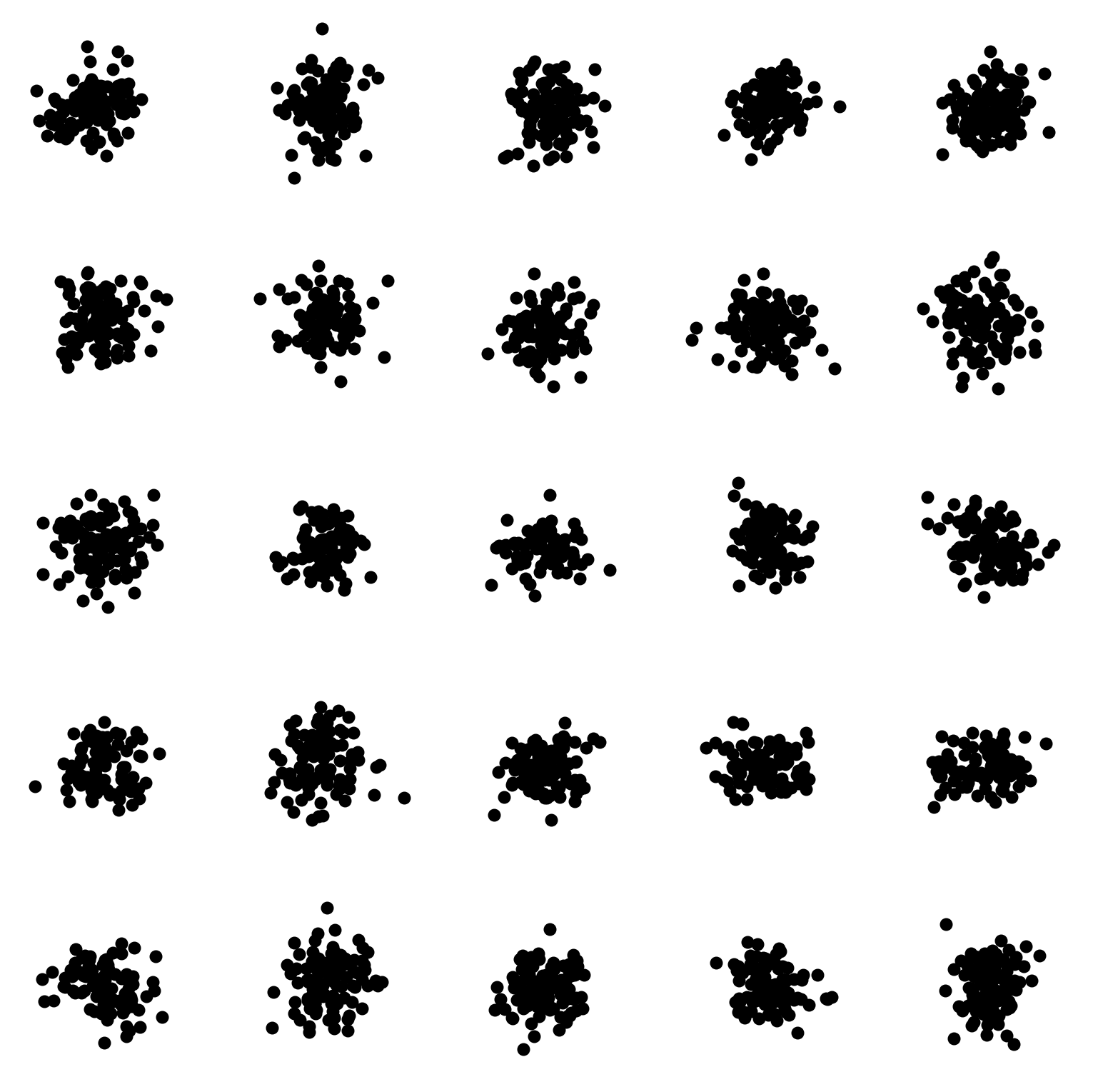}
    };

    \node[nf_plot] (iter1_unbiased) at (5,0) {
    \includegraphics[width=0.2\linewidth]
    {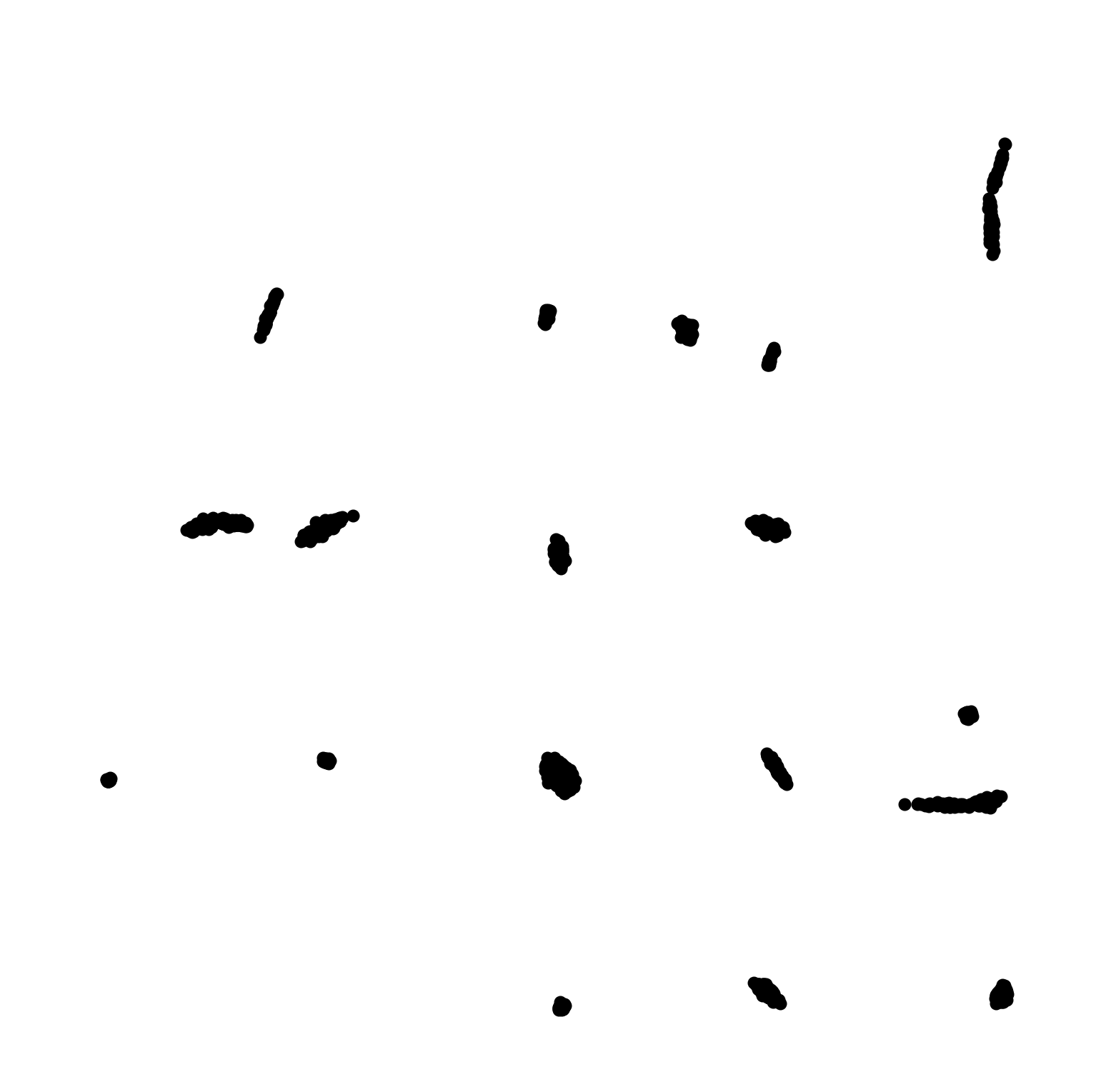}
    };

    \node[nf_plot] (iter16_unbiased) at (10,0) {
    \includegraphics[width=0.2\linewidth]
    {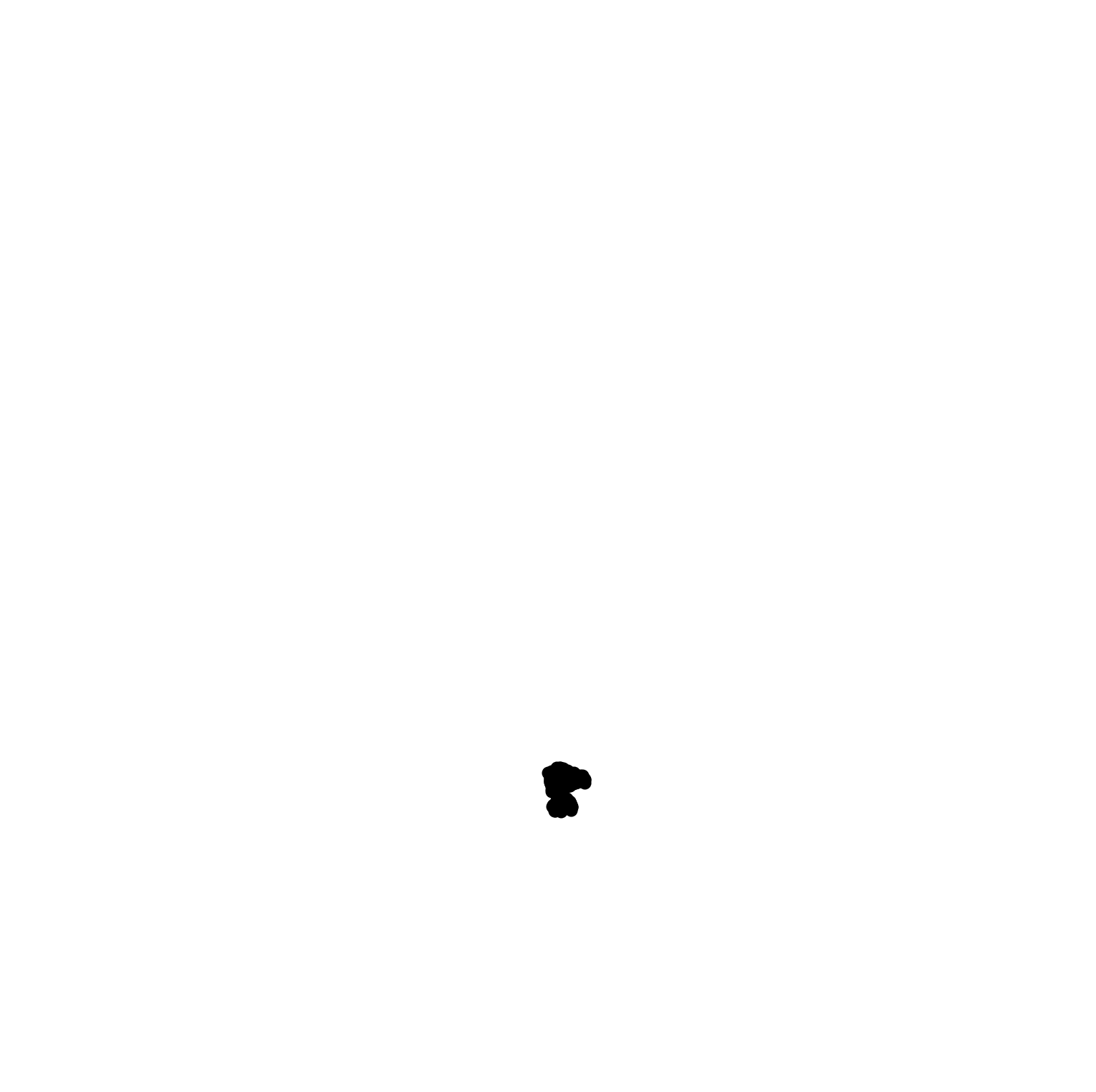}
    };

    \node[above=1mm of original] {$t=1$};
    \node[above=1mm of iter1_unbiased] {$t=200$};
    \node[above=1mm of iter16_unbiased] {$t=2k$};

    \draw[line width=2pt, -latex, postaction={decorate, decoration={raise=2ex, text along path, text align=center, text={$\bias=1$}}}]
    (original) to (iter1_unbiased);

    \draw[line width=2pt, -latex, postaction={decorate, decoration={raise=2ex, text along path, text align=center, text={$\bias=1$}}}]
    (iter1_unbiased) to (iter16_unbiased);

\end{tikzpicture}
\caption{Estimated GMM \cite{your_gan_is_secretly} distributions after $1$, $200$, and $2k$ iterations of a \synthetic{}. Notice that the modes are lost asymptotically.}
\label{fig:gmm_distributions}
\end{figure}
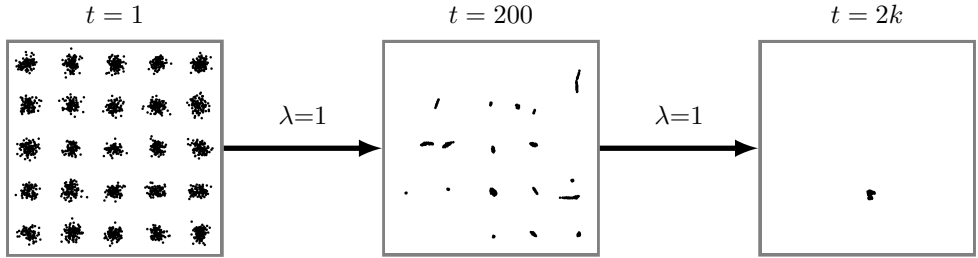

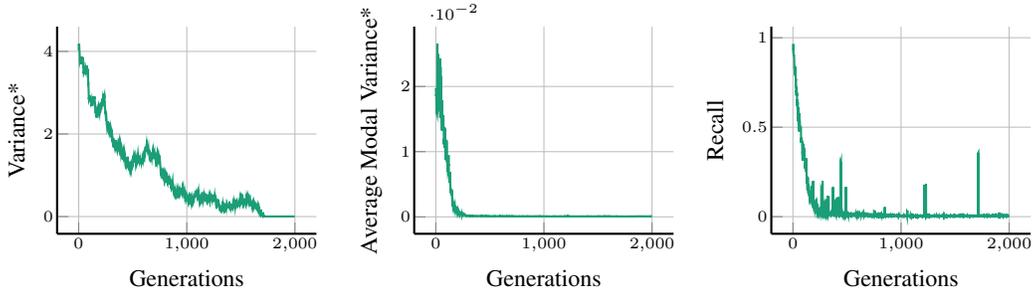
\begin{figure}[!h]

\centering
\begin{tikzpicture}
    
\pgfplotLL

\pgfplotsset{/pgfplots/group/every plot/.append style = {
    thick, no marks
}};

\begin{groupplot}[group style = {group size = 3 by 1, horizontal sep = \horizontalsep}, width = 0.36\linewidth]

    \nextgroupplot[
        title = {},
        ylabel={\small Variance*},
        xlabel={\small Generations},
        axis x line*=bottom,
        axis y line*=left,
        grid]

        \addplot
        table [
            x index=0, 
            y index=1, 
            col sep=comma] {csv/gmm/gmm_fully_synthetic.csv};

    \nextgroupplot[
        title = {},
        ylabel={\small Average Modal Variance*},
        xlabel={\small Generations},
        axis x line*=bottom,
        axis y line*=left,
        grid]

        \addplot
        table [
            x index=0, 
            y index=2, 
            col sep=comma] {csv/gmm/gmm_fully_synthetic.csv};

    \nextgroupplot[
        title = {},
        ylabel={\small Recall},
        xlabel={\small Generations},
        axis x line*=bottom,
        axis y line*=left,
        grid]
        
        \addplot
        table [
            x index=0, 
            y index=3, 
            col sep=comma] {csv/gmm/gmm_fully_synthetic.csv};

\end{groupplot} 

\end{tikzpicture}

\caption{For GMMs in a \synthetic{} (\cref{fig:gmm_distributions}), there are three primary potential metrics of diversity: variance*, average modal variance* (the average variance of each mode), and recall \cite{improved_precision_recall}. We observe that the overall variance (left) does not reflect the loss of modes that we see in \cref{fig:gmm_distributions} as smoothly as recall (right) and average modal variance (middle). Recall is therefore a suitable choice for measuring diversity in multimodal datasets and, unlike average modal variance, is compatible with distributions where the number of modes is not tractable (e.g., natural images). *For multidimensional datasets, we calculate variance as the trace of covariance.}
\label{fig:gmm_metrics}
\end{figure}

We also trained 2D GMMs in an unbiased \synthetic{} using the same 25-mode distribution as \cite{your_gan_is_secretly}. In \cref{fig:gmm_distributions} we see that the \synthetic{} gradually reduces the number of modes covered by the synthetic distribution. Various metrics could measure this loss in diversity, so in \cref{fig:gmm_metrics} we explore how well each metric reflects the dynamics of the \synthetic{}, finding that recall is best-equipped to measure diversity in multimodal datasets.

\subsection{Additional MNIST DDPM \synthetic{} results}

\begin{figure}[!h]

\centering
\begin{tikzpicture}
    
\pgfplotLL

\pgfplotsset{/pgfplots/group/every plot/.append style = {
    very thick, no marks,
}};

\begin{groupplot}[group style = {group size = 3 by 1, horizontal sep = \horizontalsep}, width = 0.36\linewidth]

    \nextgroupplot[
        title = {},
        xmax=10,
        ylabel={\small FID},
        xlabel={\small Generations},
        axis x line*=bottom,
        axis y line*=left,
        grid,
        legend style = { nodes={scale=0.7, transform shape},column sep = 10pt, legend columns = -1, legend to name = grouplegend5, text=black, cells={align=left},}]

        \addlegendimage{empty legend}
        \addlegendentry{DDPMs on an MNIST \synthetic{}:}
        \addplot
        table [
            x index=0, 
            y index=1, 
            col sep=comma] {csv/mnist_diff/MNIST_full_fid.csv};

        \addplot
        table [
            x index=0, 
            y index=2, 
            col sep=comma] {csv/mnist_diff/MNIST_full_fid.csv};

        \addplot
        table [
            x index=0, 
            y index=3, 
            col sep=comma] {csv/mnist_diff/MNIST_full_fid.csv};

        \addplot
        table [
            x index=0, 
            y index=4, 
            col sep=comma] {csv/mnist_diff/MNIST_full_fid.csv};

        \addlegendentryexpanded{$\bias = 1$};
        \addlegendentryexpanded{$\bias = 0.8$};
        \addlegendentryexpanded{$\bias = 0.66$};
        \addlegendentryexpanded{$\bias = 0.5$};

    \nextgroupplot[
        title = {},
        xmax=10,
        ylabel={\small Precision},
        xlabel={\small Generations},
        axis x line*=bottom,
        axis y line*=left,
        grid]

        \addplot
        table [
            x index=0, 
            y index=1, 
            col sep=comma] {csv/mnist_diff/MNIST_full_precision.csv};

        \addplot
        table [
            x index=0, 
            y index=2, 
            col sep=comma] {csv/mnist_diff/MNIST_full_precision.csv};

        \addplot
        table [
            x index=0, 
            y index=3, 
            col sep=comma] {csv/mnist_diff/MNIST_full_precision.csv};

        \addplot
        table [
            x index=0, 
            y index=4, 
            col sep=comma] {csv/mnist_diff/MNIST_full_precision.csv};

    \nextgroupplot[
        title = {},
        xmax=10,
        ylabel={\small Recall},
        xlabel={\small Generations},
        axis x line*=bottom,
        axis y line*=left,
        grid]
        
        \addplot
        table [
            x index=0, 
            y index=1, 
            col sep=comma] {csv/mnist_diff/MNIST_full_recall.csv};

        \addplot
        table [
            x index=0, 
            y index=2, 
            col sep=comma] {csv/mnist_diff/MNIST_full_recall.csv};

        \addplot
        table [
            x index=0, 
            y index=3, 
            col sep=comma] {csv/mnist_diff/MNIST_full_recall.csv};

        \addplot
        table [
            x index=0, 
            y index=4, 
            col sep=comma] {csv/mnist_diff/MNIST_full_recall.csv};

\end{groupplot}
\node at ($(group c2r1) + (0,-85pt)$) {\ref{grouplegend5}}; 

\end{tikzpicture}

\caption{The FID (left), quality (precision, middle), and diversity (recall, right) of synthetic images produced by DDPMs on MNIST.}
\label{fig:mnist_full_all}

\end{figure}
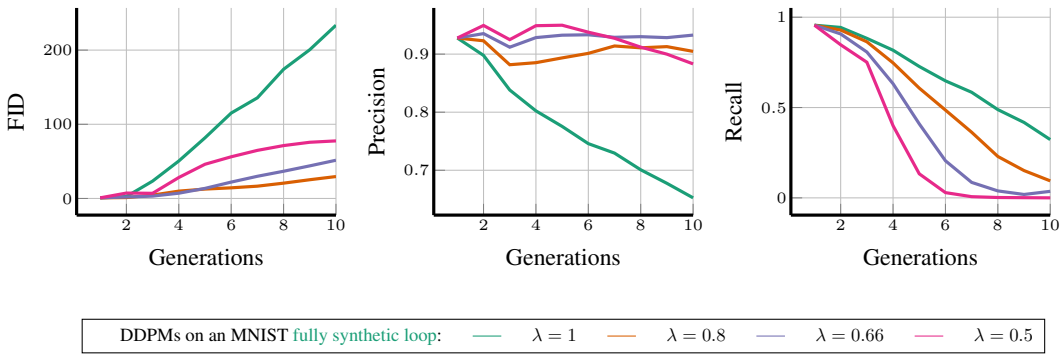

In \cref{fig:synthetic_biased_fid_precision_recall} we showcased the results of training MNIST DDPMs in a \synthetic{} with various sampling bias factors $\bias$. In \cref{fig:mnist_full_all} we have the results (FID, precision, and recall) more generations $t$ and different sampling biases $\bias$.

\subsection{Normalizing flow \synthetic{}}

We implemented the \synthetic{} using normalizing flows \cite{dinh2016density, kingma2018glow} for generative modeling of the two-dimensional Rosenbrock reference distribution \cite{rosenbrock} in order to visualize the outcome of this particular scenario in a controlled setting. Normalizing flows are unique in that they enable exact evaluation of the likelihood of the estimated distribution due to their invertibility \cite{dinh2016density}. This leads to a relatively straightforward training procedure compared to GANs, which often require careful balancing between the generator and discriminator networks to avoid mode collapse \cite{NIPS2017_892c3b1c}. Therefore, by using a low-dimensional reference distribution, this setup allows us to demonstrate the \synthetic{} while eliminating potential training imperfections.

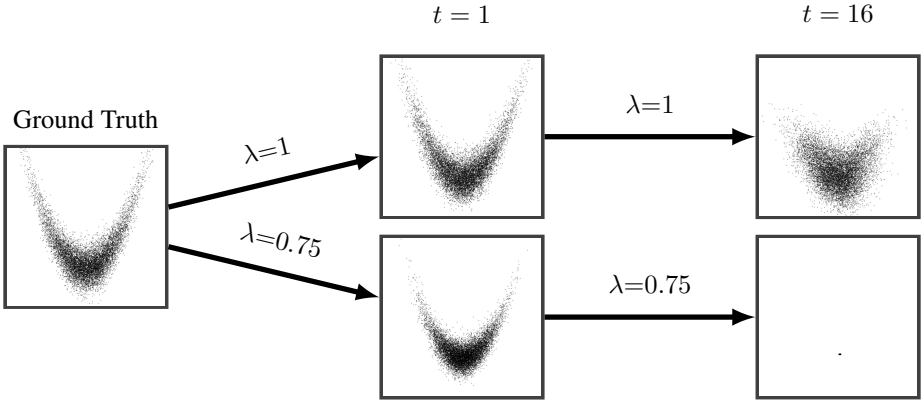
\begin{figure}[!h]
\centering

\begin{tikzpicture}

    \tikzstyle{nf_plot} = [draw, darkgray, line width=1mm, inner sep=0pt]

    \node[nf_plot] (original) {
    \includegraphics[width=0.15\linewidth]
    {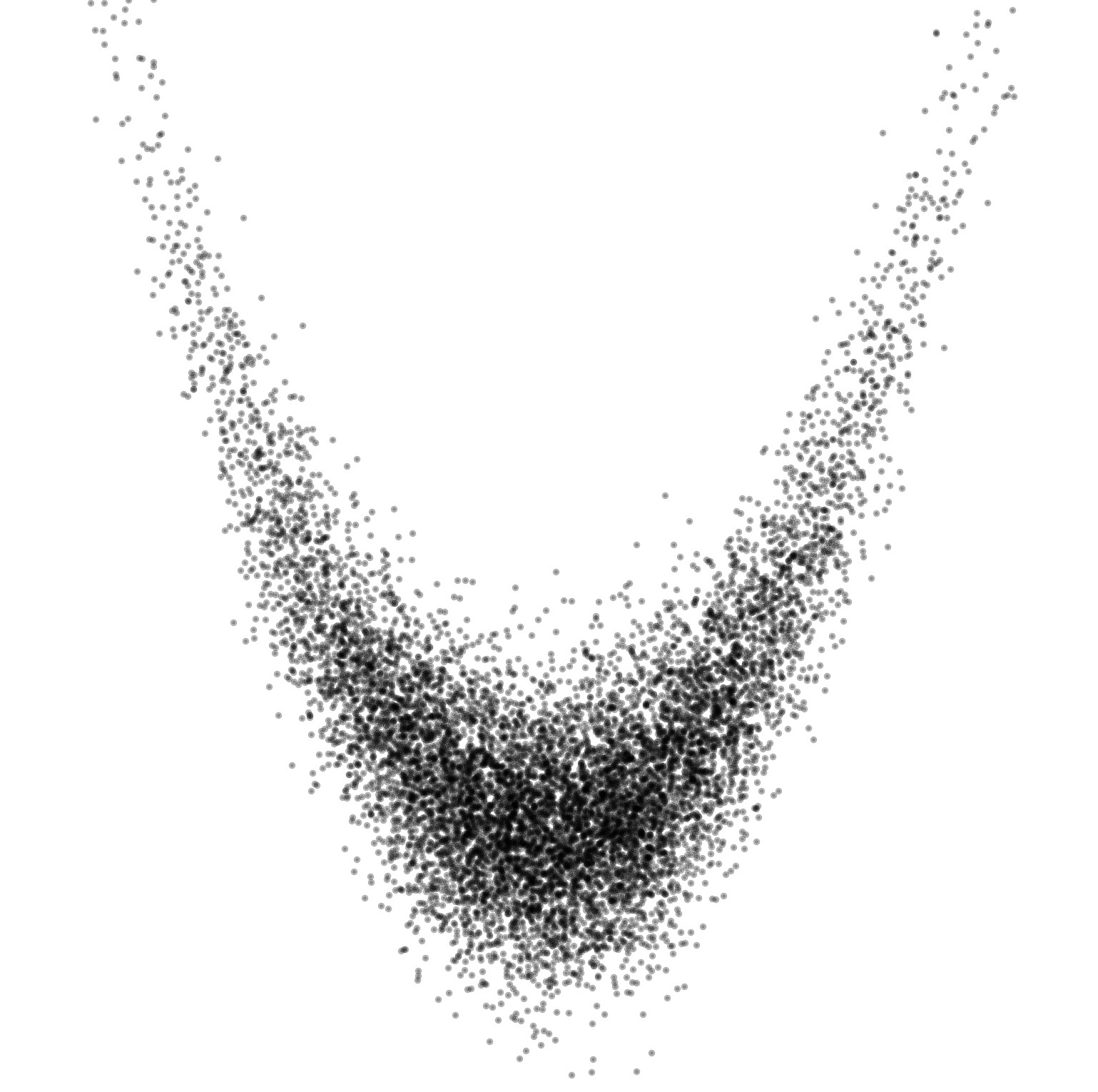}
    };

    \node[nf_plot] (iter1_unbiased) at (5,1.2) {
    \includegraphics[width=0.15\linewidth]
    {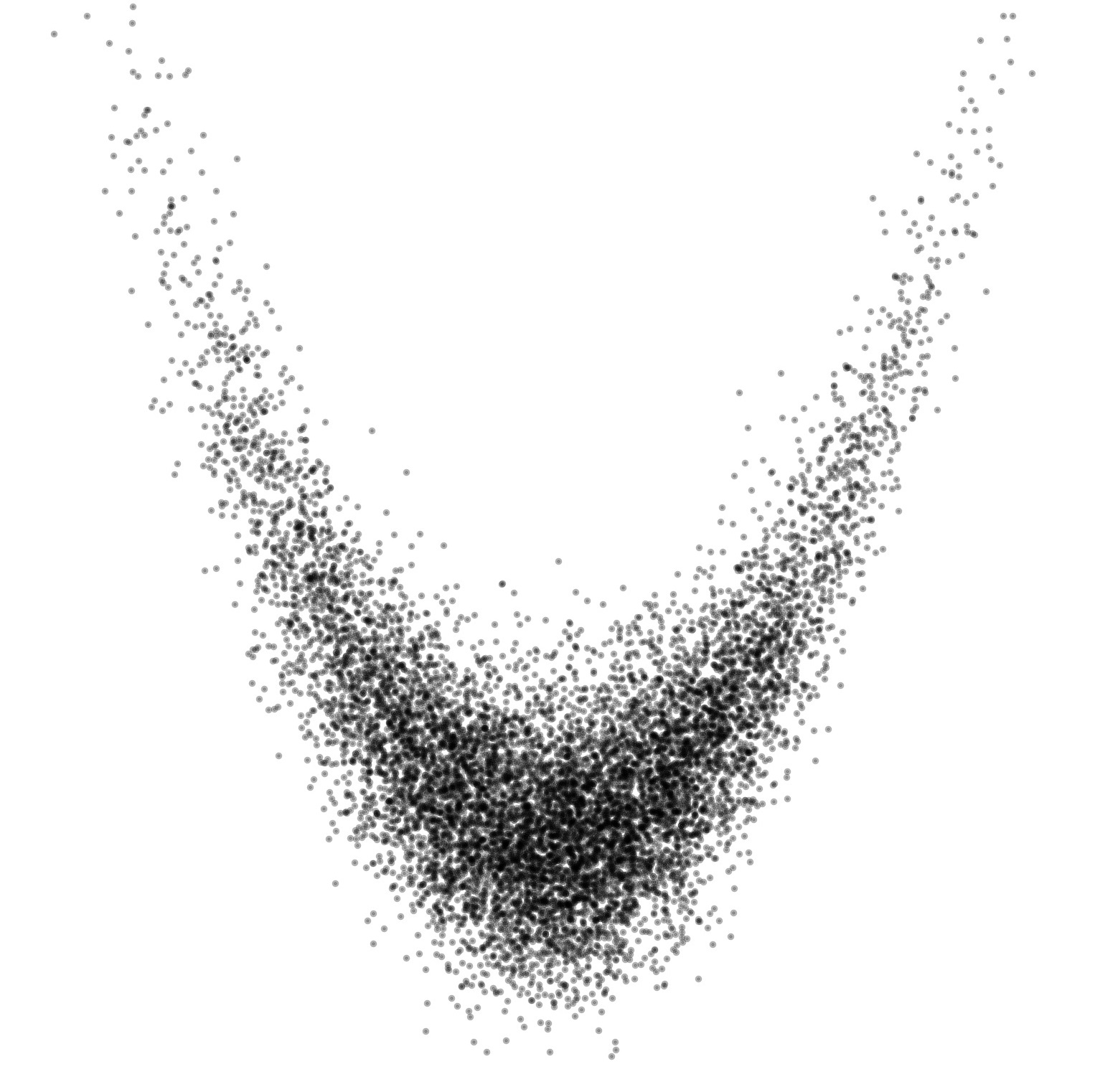}
    };

    \node[nf_plot] (iter1_biased) at (5,-1.2) {
    \includegraphics[width=0.15\linewidth]
    {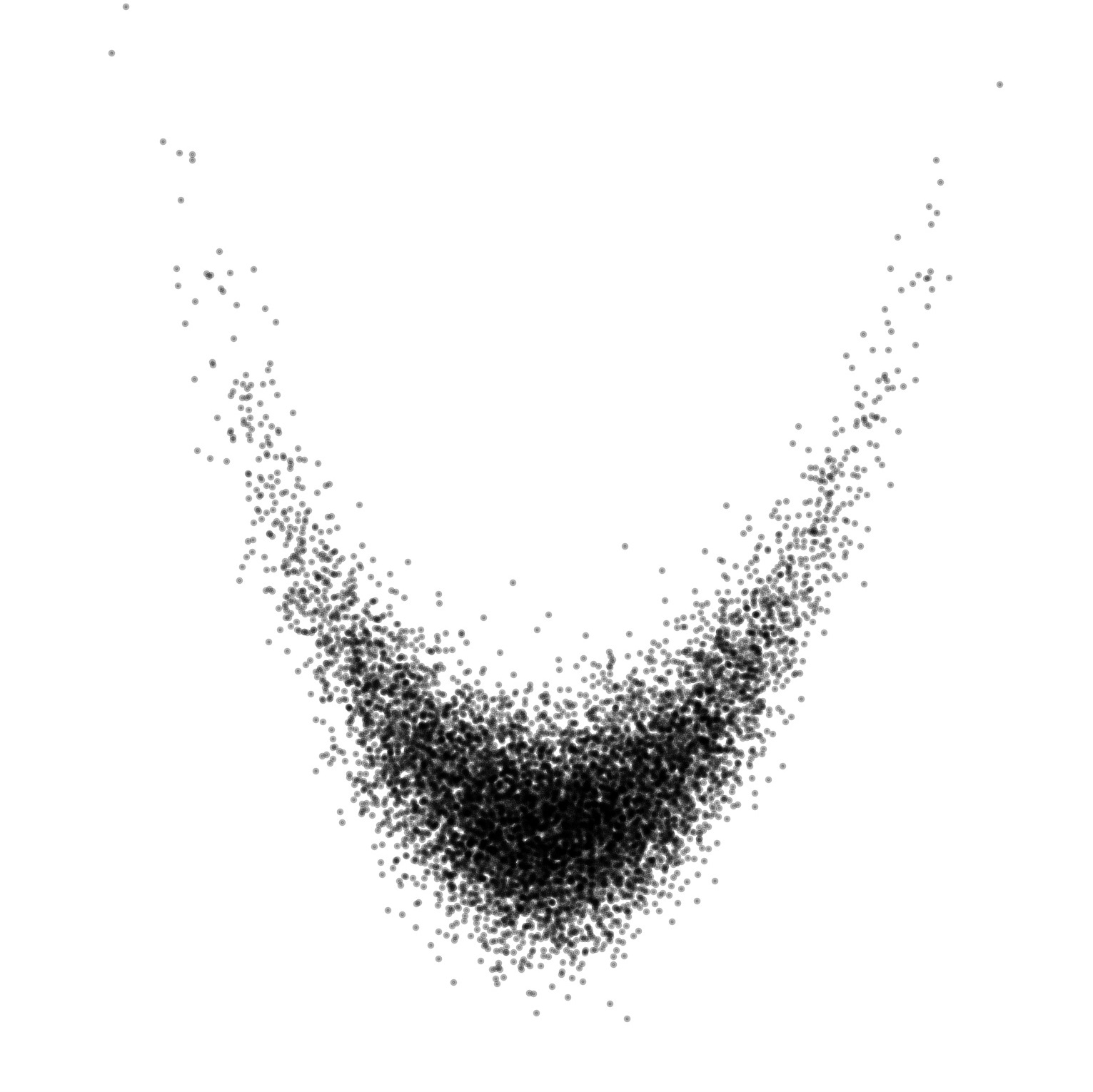}
    };

    \node[nf_plot] (iter16_unbiased) at (10,1.2) {
    \includegraphics[width=0.15\linewidth]
    {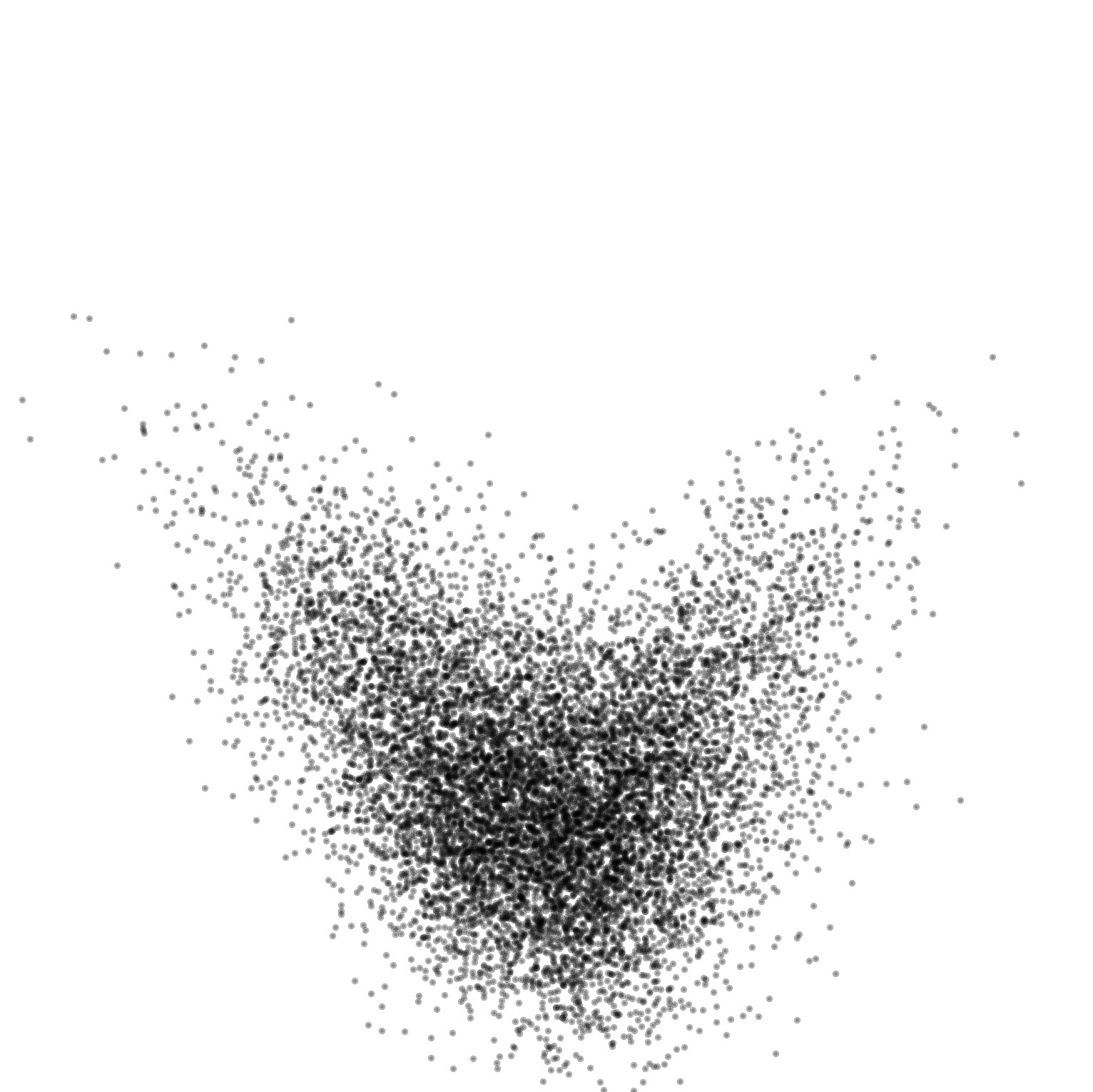}
    };

    \node[nf_plot] (iter16_biased) at (10,-1.2) {
    \includegraphics[width=0.15\linewidth]
    {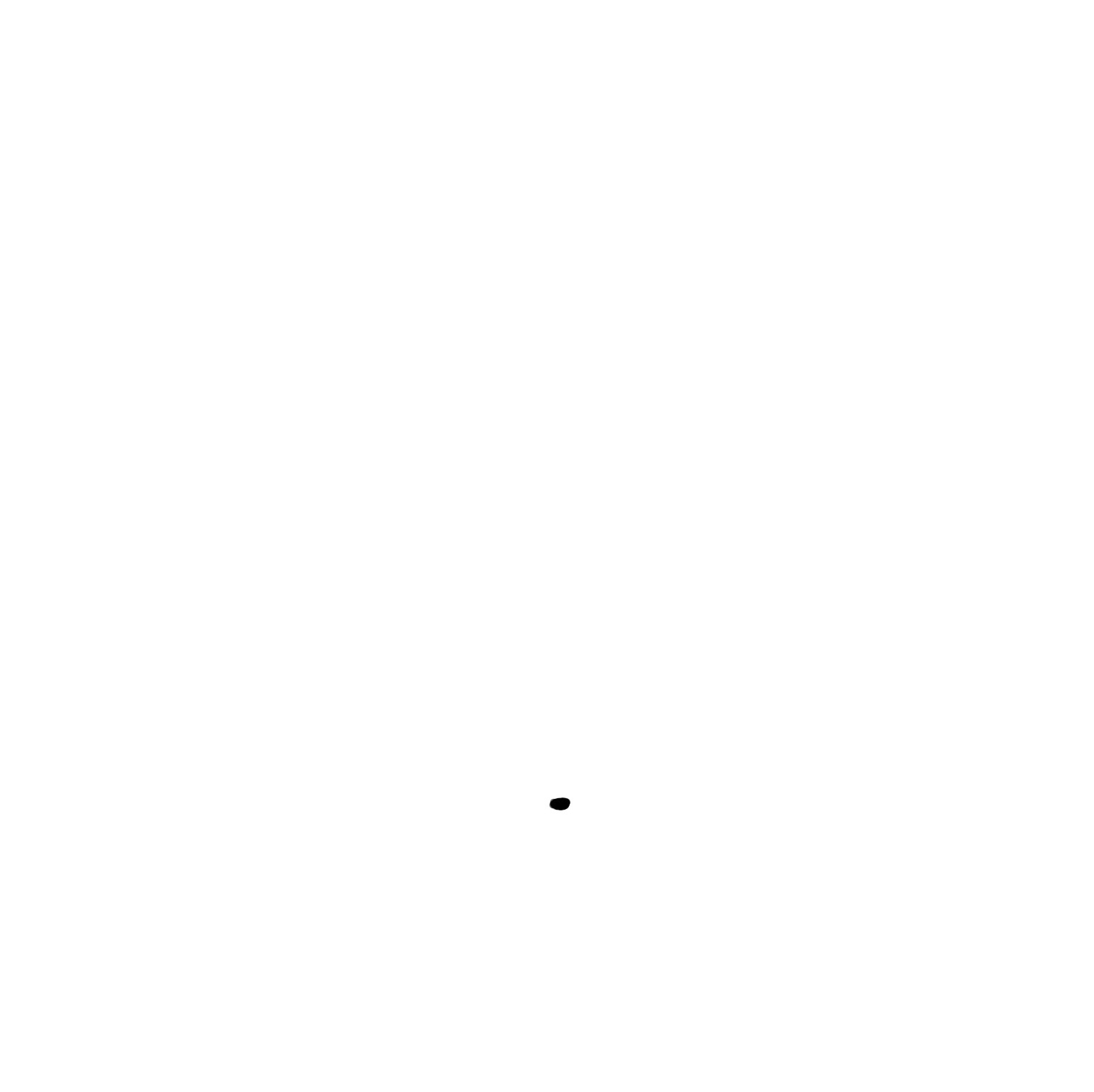}
    };

    \node[above=1mm of original] {Ground Truth};
    \node[above=3mm of iter1_unbiased] {$t=1$};
    \node[above=3mm of iter16_unbiased] {$t=16$};

    \draw[line width=2pt, -latex, postaction={decorate, decoration={raise=2ex, text along path, text align=center, text={$\bias=1$}}}]
    (original) to (iter1_unbiased);

    \draw[line width=2pt, -latex, postaction={decorate, decoration={raise=2ex, text along path, text align=center, text={$\bias=0.75$}}}]
    (original) to (iter1_biased);

    \draw[line width=2pt, -latex, postaction={decorate, decoration={raise=2ex, text along path, text align=center, text={$\bias=1$}}}]
    (iter1_unbiased) to (iter16_unbiased);

    \draw[line width=2pt, -latex, postaction={decorate, decoration={raise=2ex, text along path, text align=center, text={$\bias=0.75$}}}]
    (iter1_biased) to (iter16_biased);

\end{tikzpicture}
\caption{The \synthetic{} implememted with a formalizing flow \cite{dinh2016density} applied to the 2D Rosenbrock distribution \cite{rosenbrock}. Sampling with or without bias still loses the tails of the distribution (i.e., diversity). Using $\bias<1$ accelerates this loss of diversity.}
\label{fig:nf}
\end{figure}

According to the \synthetic{} setup, we start with a training dataset of $10^4$ samples from the 2D Rosenbrock distribution with the density function $\mathcal{P}_r(x_1, x_2) \propto \exp \left( - \frac{1}{2} x_{1}^{2} - \left(x_{2}-x_{1}^{2} \right)^{2}\right)$ \cite{rosenbrock}, which is plotted on the left-hand side of \Cref{fig:nf}. The subsequent generations of normalizing flow models are trained using synthetic data generated by the previous pre-trained normalizing flow for $16$ generations, both with and without sampling bias. We employ the GLOW normalizing flow architecture \cite{kingma2018glow} with eight coupling layers \cite{kingma2018glow} and a hidden dimension of $64$. The training is carried out for $20$ epochs with a batch size of $256$ for each generation, ensuring convergence as determined by monitoring the model's likelihood over a validation set. \Cref{fig:nf} summarizes the results of this \synthetic{} setup. To incorporate sampling bias, we sample from $\mathcal{N}(\bm{0}_d, \bias\mI_d)$ from the latent space of the model, where $d = 2$. As shown, regardless of the presence of sampling bias, the resulting distribution after $16$ generations loses the tails of the reference distribution, indicating a loss of diversity. This phenomenon becomes more pronounced when sampling bias is present ($\bias<1$).

\clearpage
\section{FFHQ \synthetic{} images with $\bias=1$}
\label{sec:ffhq_samples_unbiased}

We show additional randomly chosen synthetic samples produced by the same StyleGAN FFHQ unbiased \synthetic{} as in \cref{fig:ffhq_samples} and \cref{fig:synthetic_unbiased_fid_precision_recall}.

\begin{figure}[!h]
    \centering
    \includegraphics[width=\linewidth]{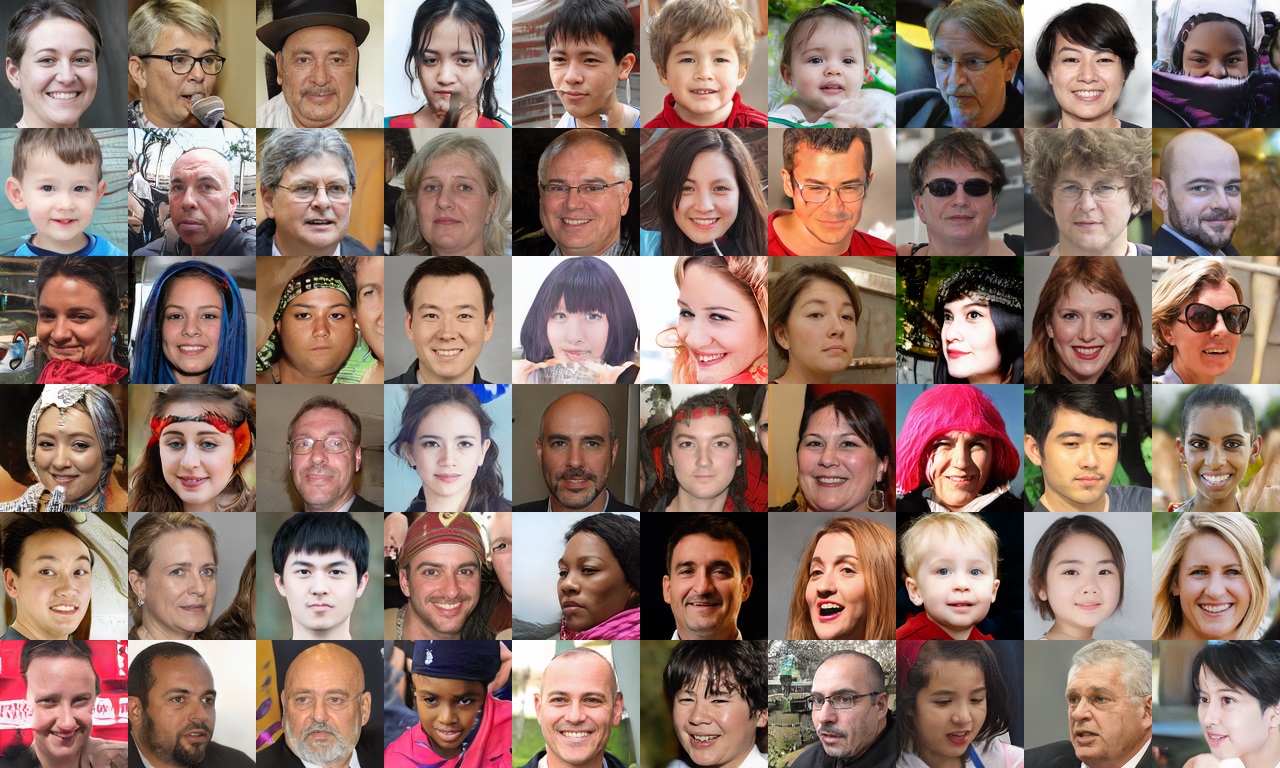}
    \caption{Generation $t=1$ of a \synthetic{} with bias $\lambda=1$. i.e., synthetic samples from the first model $\model^1$.}
    \label{fig:ffhq_samples_gen1_unbiased}
\end{figure}

\begin{figure}[!h]
    \centering
    \includegraphics[width=\linewidth]{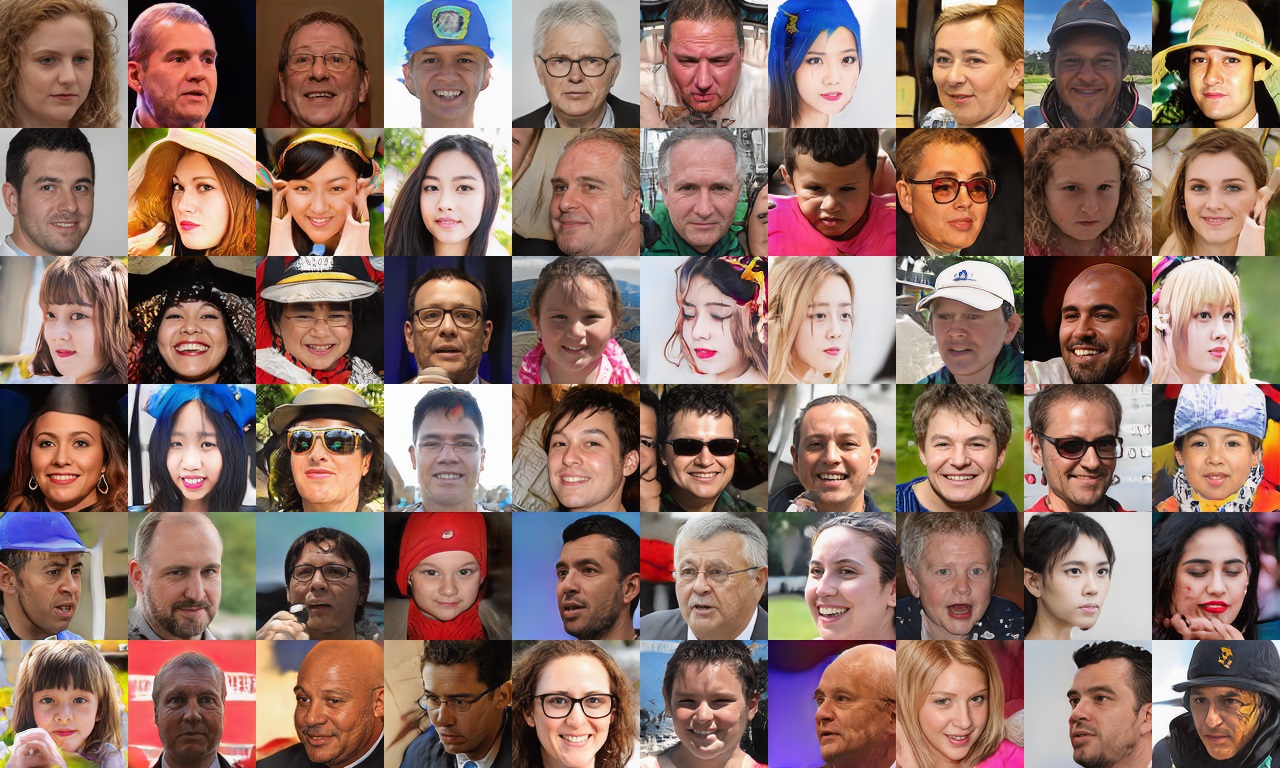}
    \caption{Generation $t=3$ of a \synthetic{} with bias $\lambda=1$}
\end{figure}

\clearpage
\begin{figure}[!h]
    \centering
    \includegraphics[width=\linewidth]{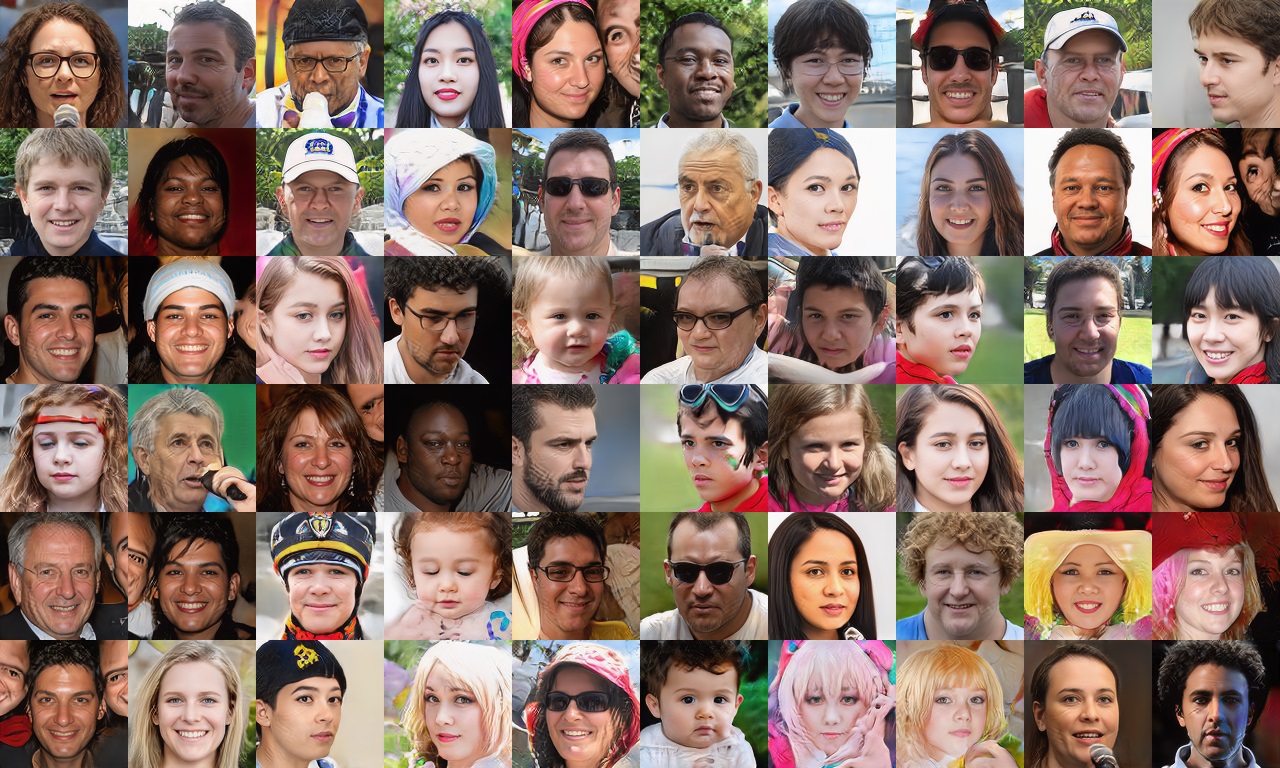}
    \caption{Generation $t=5$ of a \synthetic{} with bias $\lambda=1$}
\end{figure}

\begin{figure}[!h]
    \centering
    \includegraphics[width=\linewidth]{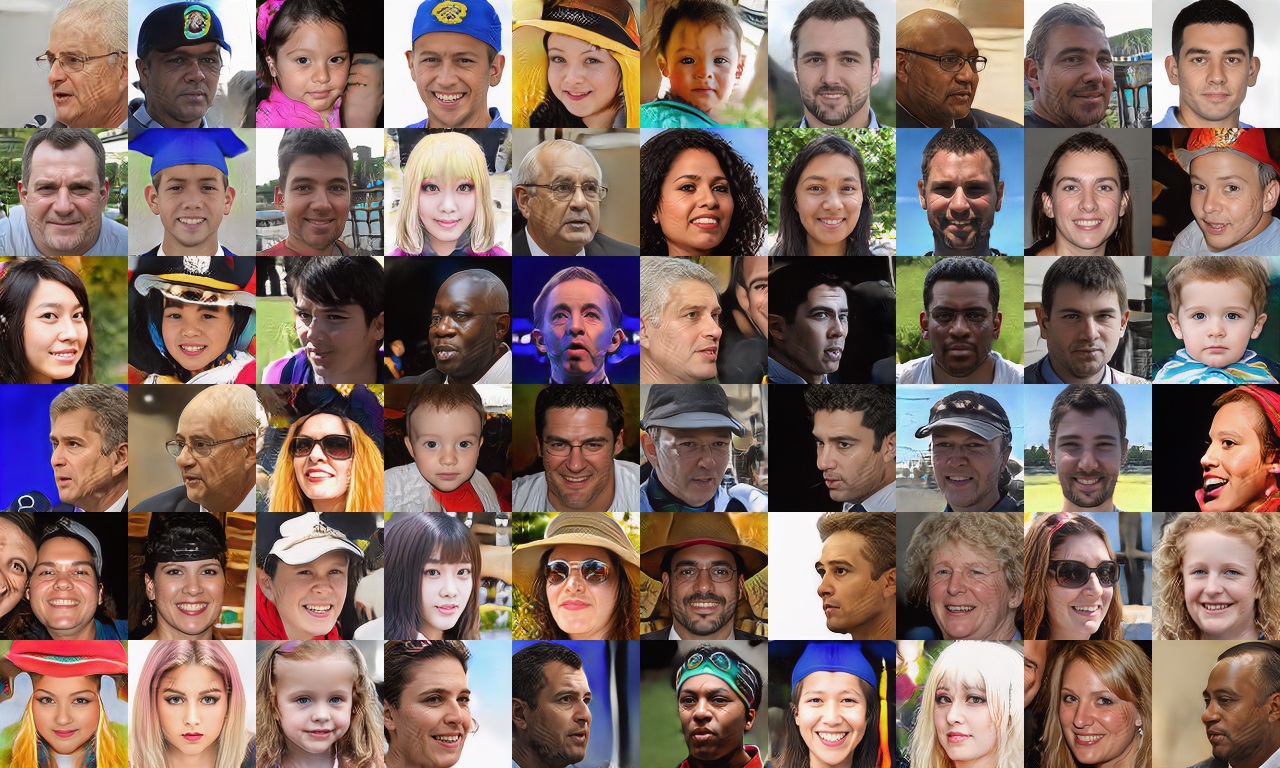}
    \caption{Generation $t=7$ of a \synthetic{} with bias $\lambda=1$}
\end{figure}

\clearpage
\begin{figure}[!h]
    \centering
    \includegraphics[width=\linewidth]{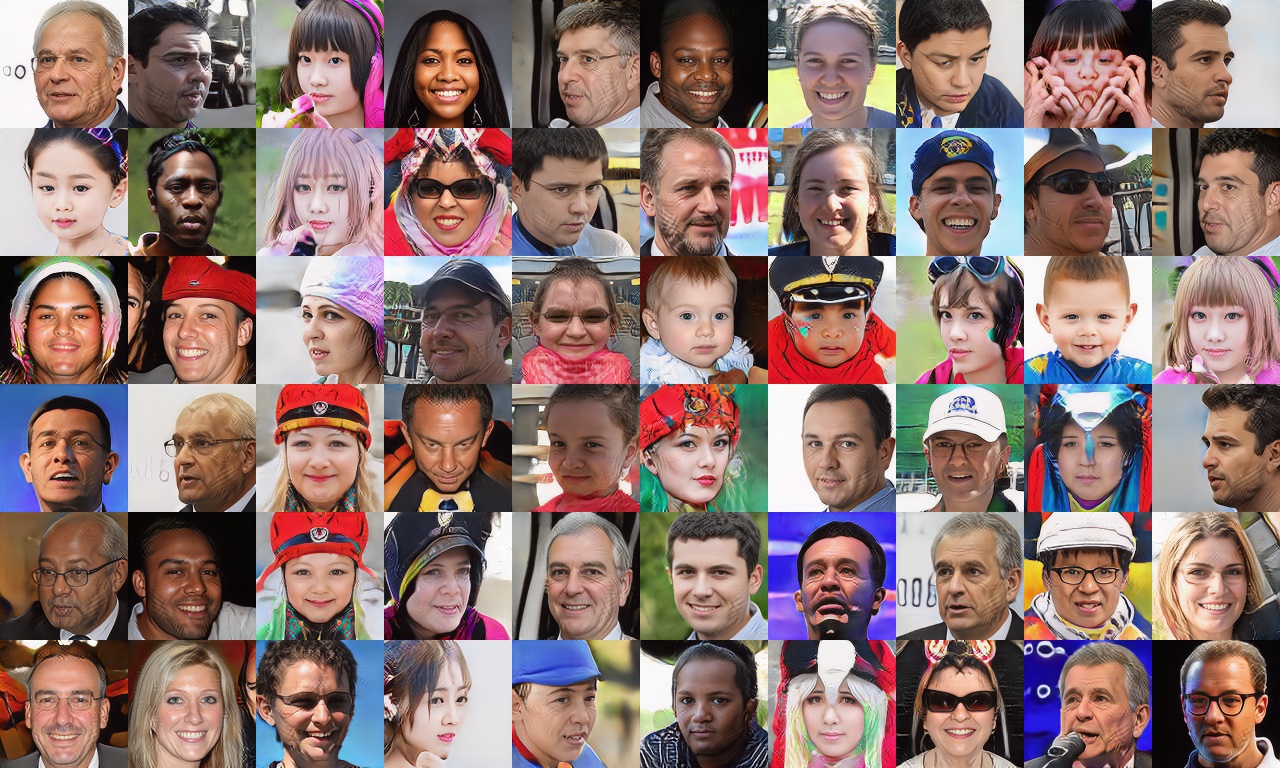}
    \caption{Generation $t=9$ of a \synthetic{} with bias $\lambda=1$}
\end{figure}

\section{FFHQ \synthetic{} images with $\bias=0.7$}
\label{sec:ffhq_samples_biased}

As in \cref{sec:ffhq_samples_unbiased}, here we show synthetic FFHQ images produced by a StyleGAN architecture in a \synthetic{} with biased sampling ($\lambda=0.7$, \cref{fig:synthetic_biased_fid_precision_recall}) that slows the proliferation of artifacts, but at the cost of severely decreased diversity.

\begin{figure}[!h]
    \centering
    \includegraphics[width=\linewidth]{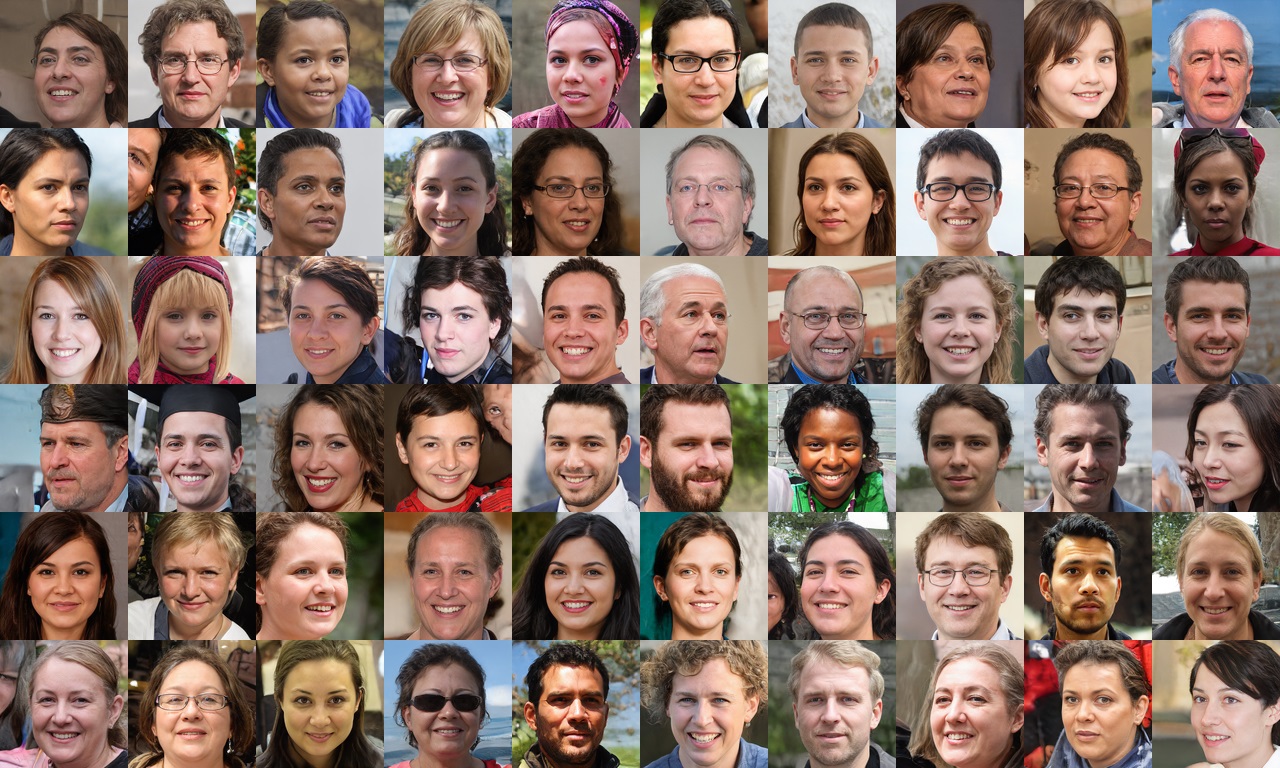}
    \caption{Generation $t=1$ of a \synthetic{} with bias $\lambda=0.7$}
\end{figure}

\clearpage
\begin{figure}[!h]
    \centering
    \includegraphics[width=\linewidth]{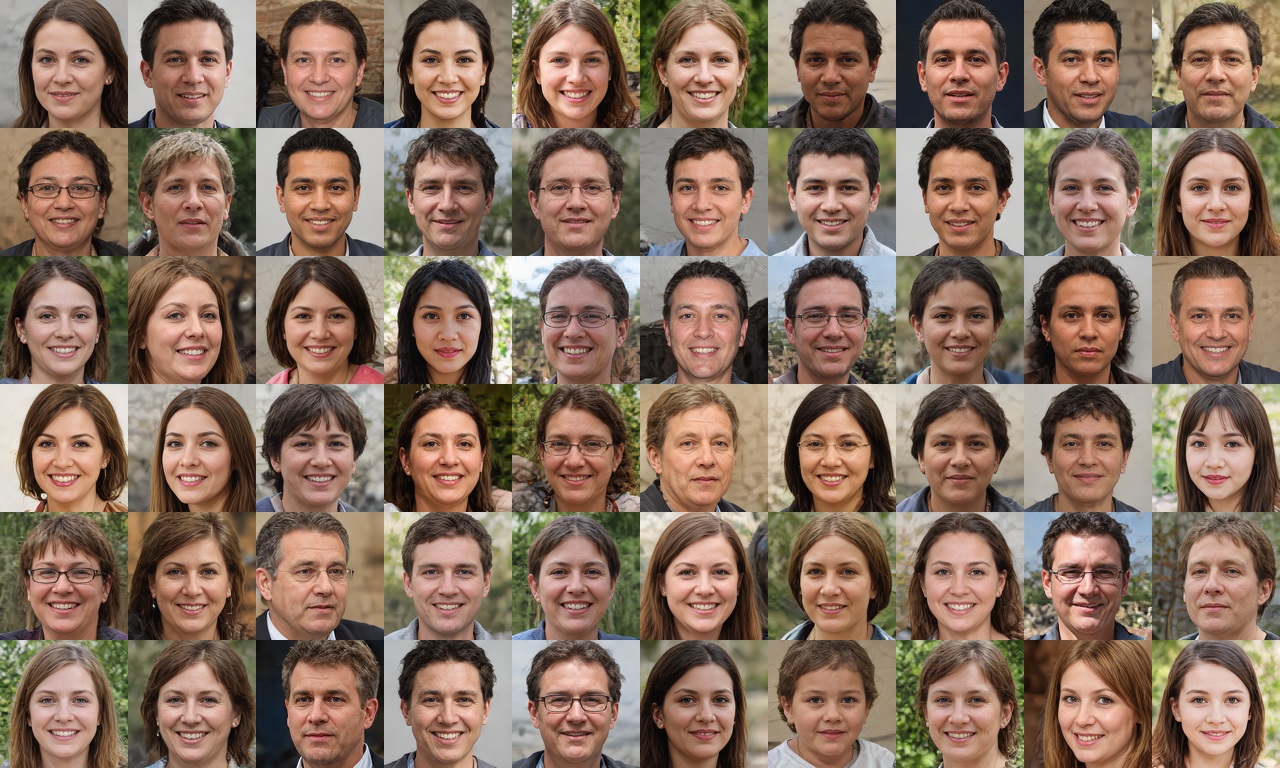}
    \caption{Generation $t=3$ of a \synthetic{} with bias $\lambda=0.7$}
\end{figure}

\begin{figure}[!h]
    \centering
    \includegraphics[width=\linewidth]{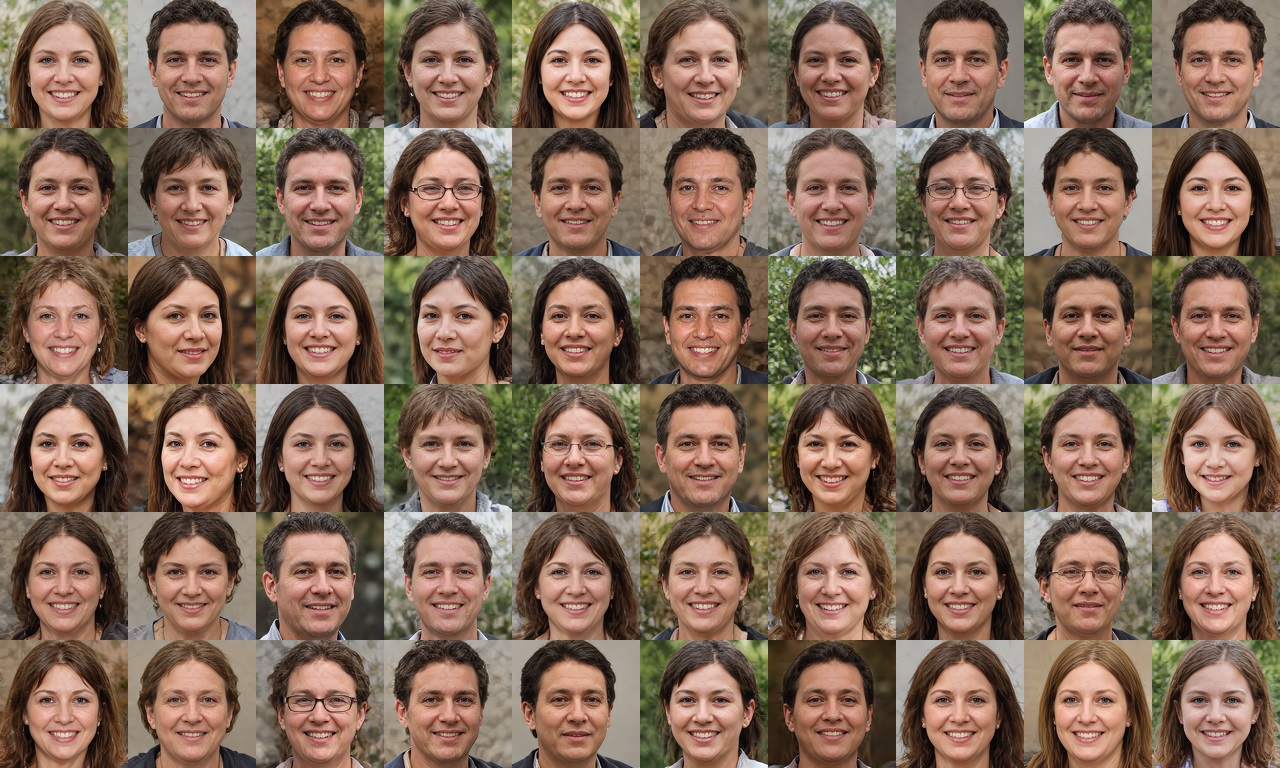}
    \caption{Generation $t=5$ of a \synthetic{} with bias $\lambda=0.7$}
\end{figure}

\clearpage
\section{MNIST \synthetic{} images}

Here we show randomly chosen samples from each generation of an MNIST DDPM in a \synthetic{} for different sampling biases (as discussed in \cref{fig:synthetic_unbiased_fid_precision_recall} and \cref{fig:synthetic_biased_fid_precision_recall}).

\newcommand{\qq}{\hspace*{4mm}}
\newcommand{\qqq}{\hspace*{3mm}}

\begin{figure}[!h]
    \hspace*{-5mm}\small Gen.\qqq\foreach \b in {1,2,3,4,5,6,7,8,9,10,11,12,13,14,15,16,17,18,19,20}{
    \begin{minipage}{0.0365\linewidth}
    \small \b
    \end{minipage}
    }
    
    \includegraphics[width=\linewidth]{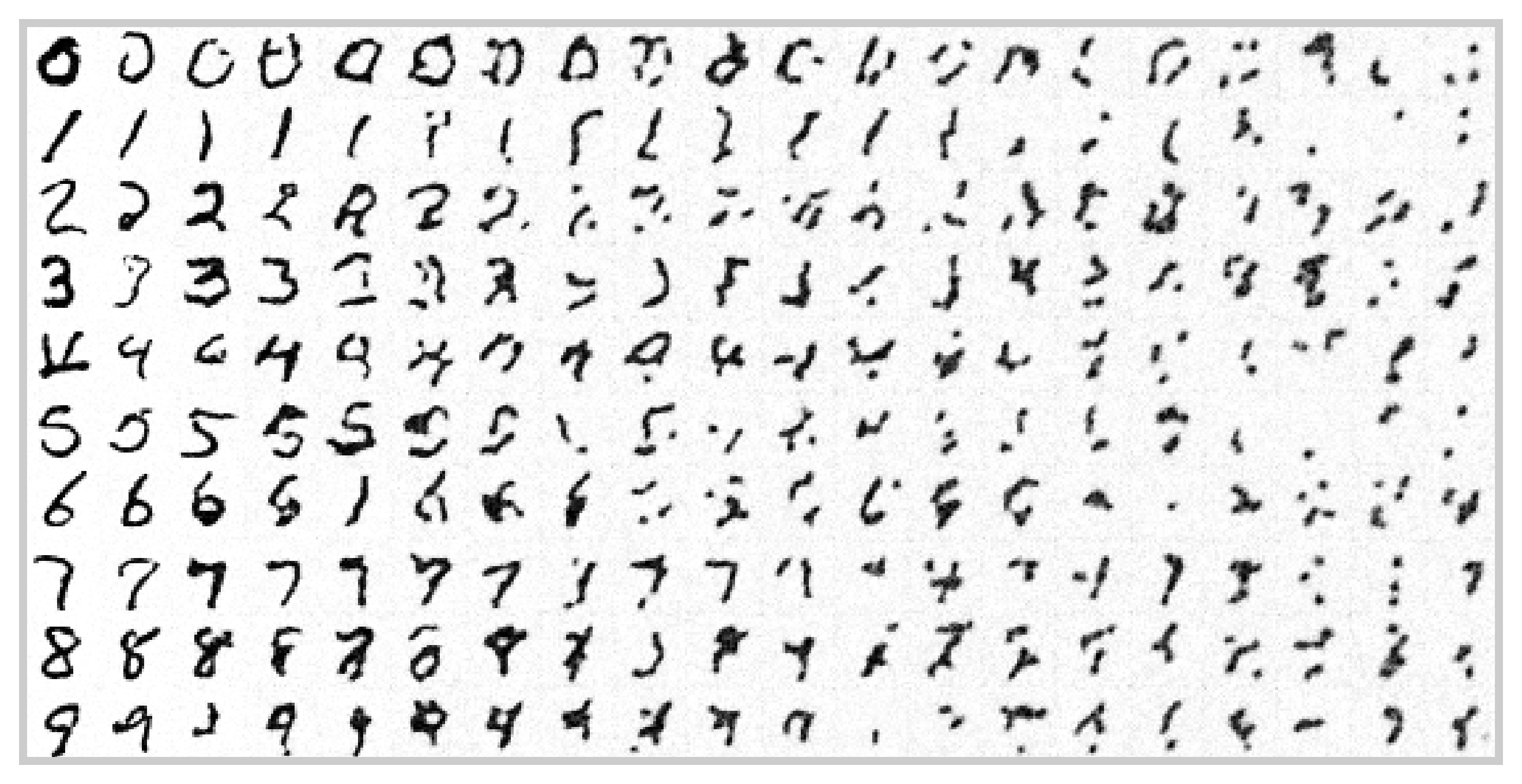}\\
  
\caption{\textbf{Without sampling bias, synthetic data modes drift from real modes and merge together.} Randomly selected synthetic MNIST images of each generation without sampling bias ($\bias = 1$).}

\label{fig:mnist_img_nobias}
\end{figure}

\begin{figure}[!h]
    \hspace*{-5mm}\small Gen.\qqq\foreach \b in {1,2,3,4,5,6,7,8,9,10,11,12,13,14,15,16,17,18,19,20}{
    \begin{minipage}{0.0365\linewidth}
    \small \b
    \end{minipage}
    }
    
    \includegraphics[width=\linewidth]{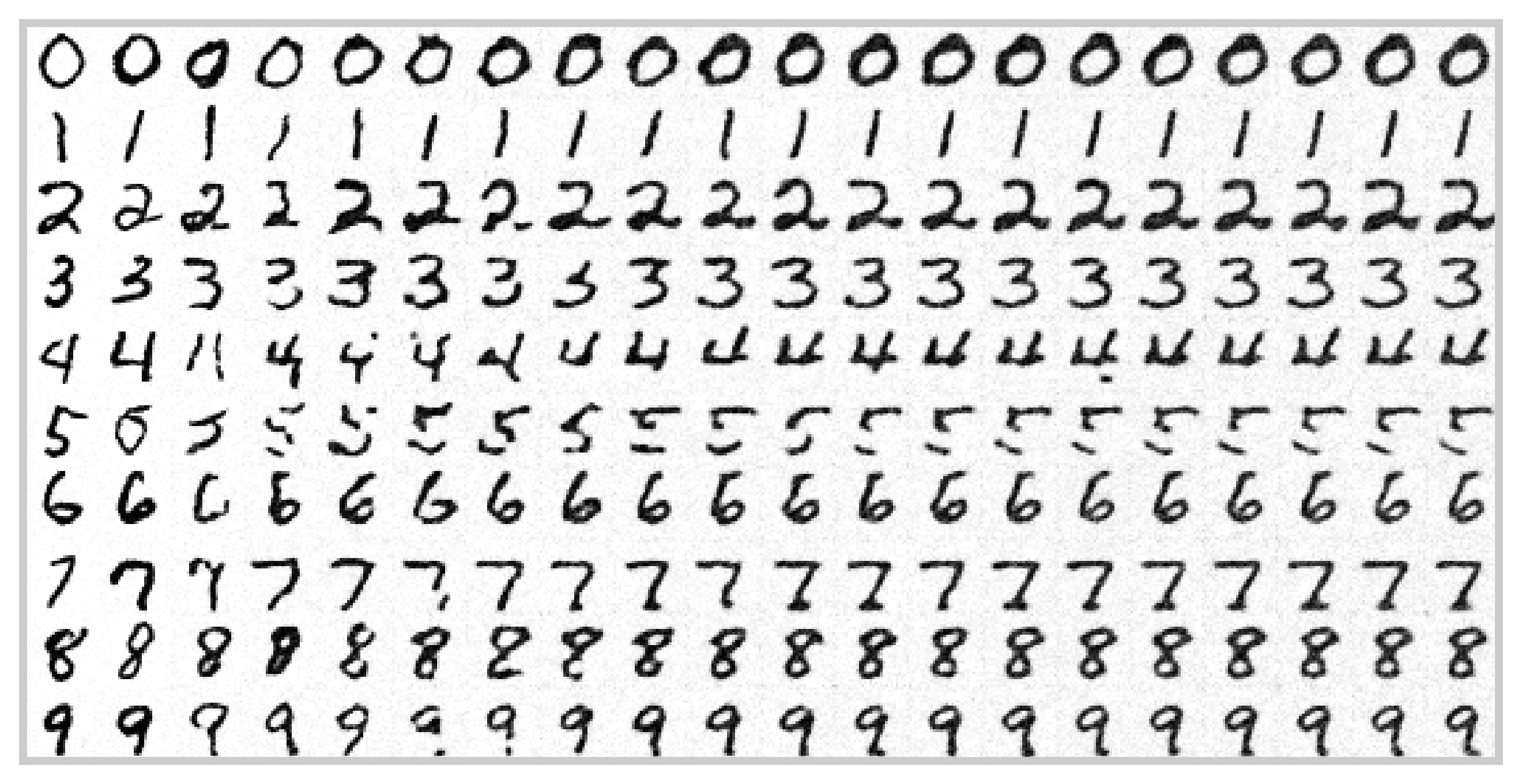}\\
  
\caption{\textbf{With sampling bias, synthetic data modes drift and collapse around a single (high quality) image before merging.} Randomly selected synthetic MNIST images of each generation without sampling bias ($\bias = 0.8$).}

\label{fig:mnist_img_withbias}
\end{figure}

\clearpage
\section{FFHQ \mixed{} images with $\bias=1$}
\label{sec:ffhq_samples_mixed_unbiased}

\begin{figure}[!h]
    \centering
    \includegraphics[width=\linewidth]{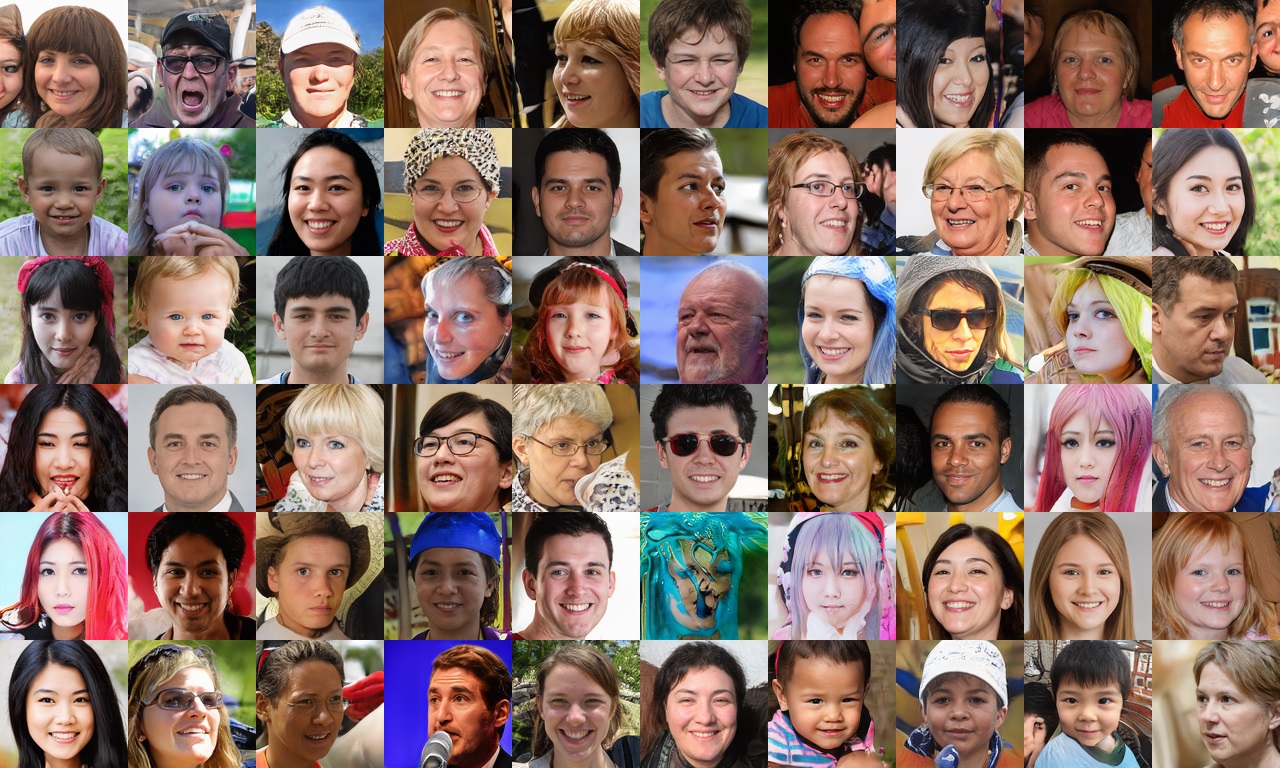}
    \caption{Generation $t=3$ of a \mixed{} with bias $\lambda=1$. See \cref{fig:ffhq_samples_gen1_unbiased} for the samples from $t=1$ (in any autophagous loop the first model $\model^1$ always trains on purely real data, see \cref{sec:background}).}
\end{figure}

\begin{figure}[!h]
    \centering
    \includegraphics[width=\linewidth]{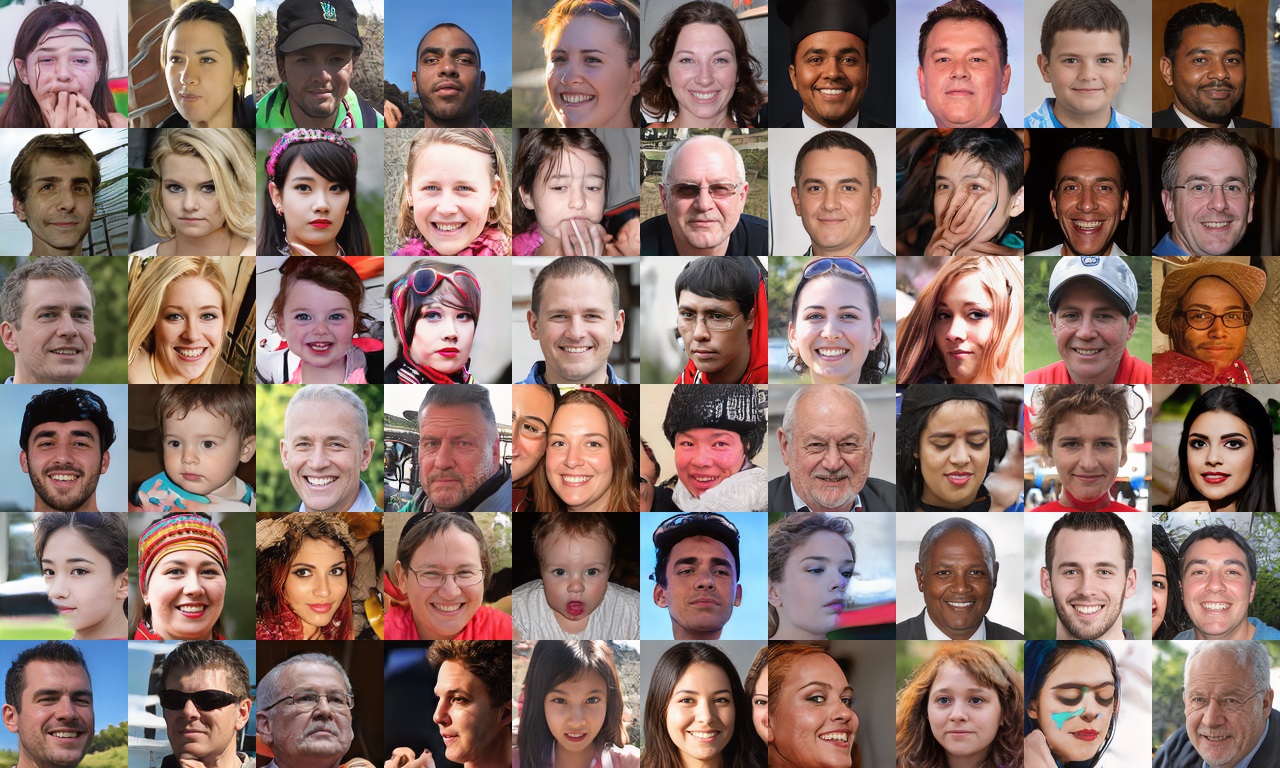}
    \caption{Generation $t=6$ of a \mixed{} with bias $\lambda=1$}
\end{figure}



    







\clearpage
\section{Additional results for the \fresh{} }
\label{freshappendix}

Here we provide three additional Gaussian experiments investigating the convergence of the \fresh{}. 

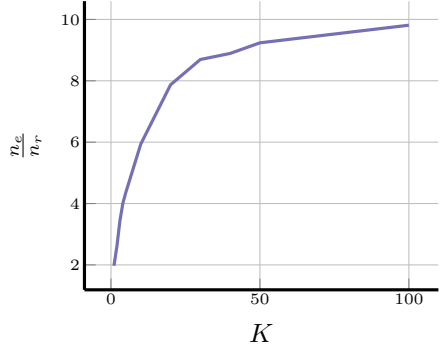
\begin{figure}[!bh]

\centering
\begin{tikzpicture}
    
\pgfplotLL

\pgfplotsset{/pgfplots/group/every plot/.append style = {
    very thick, no marks,
}};

\begin{groupplot}[group style = {group size = 3 by 1, horizontal sep = \horizontalsep}, width = 0.45\linewidth]

    \nextgroupplot[
        title = {},
        ylabel={$\frac{n_e}{n_r}$},
        xlabel={$K$},
        axis x line*=bottom,
        axis y line*=left,
        grid,
        legend style={
                at={(0.05,.25)},
                anchor=west,
                font=\footnotesize},
                legend style={nodes={scale=0.65, transform shape}}
        ]

        \addplot[\colorFresh]
        table [
            x index= 0, 
            y index= 5, 
            col sep=comma] {csv/fresh/differentK.csv};
        ]

\end{groupplot}

\end{tikzpicture}

\caption{ The effective sample size $n_e$ divided by real sample size $n_r$ for different numbers of accessed previous generations $K$.}
\label{fig:fresh_appendix_2}

\end{figure}

\textbf{Experiment 1}: In \cref{section5} we assumed that we only sample from the previous generation $\model^{t-1}$ for creating the synthetic dataset $\data_s^t$. In this experiment we sample randomly from $K$ previous models $(\model^{\tau})_{\tau = t-1-K}^{t-1}$. Here $n_r = 10^3$, $n_s = 10^4$, and $\bias = 1$. In Figure \ref{fig:fresh_appendix_2} we see how $\frac{n_e}{n_r}$ varies with respect to $K$. Increasing the memory $K$ in sampling from previous generations can boost performance, however the rate of improvement becomes slower as $K$ increases.

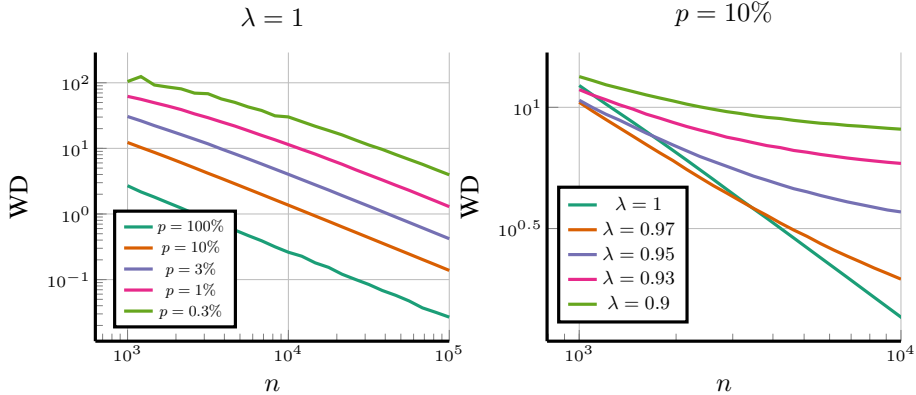
\begin{figure}[!bh]

\centering
\begin{tikzpicture}
    
\pgfplotLL

\pgfplotsset{/pgfplots/group/every plot/.append style = {
    very thick, no marks,
}};

\begin{groupplot}[group style = {group size = 3 by 1, horizontal sep = \horizontalsep}, width = 0.45\linewidth]

    \nextgroupplot[
        title = {$\bias = 1$},
        xmax=100000,
        xmode = log,
        ymode = log,
        ylabel={WD},
        xlabel={$n$},
        axis x line*=bottom,
        axis y line*=left,
        grid,
        legend style={
                at={(0.05,.25)},
                anchor=west,
                font=\footnotesize},
                legend style={nodes={scale=0.65, transform shape}}
        ]

        \addplot
        table [
            x index=0, 
            y index=5, 
            col sep=comma] {csv/fresh/p.csv};

        \addplot
        table [
            x index=0, 
            y index=4, 
            col sep=comma] {csv/fresh/p.csv};

        \addplot
        table [
            x index=0, 
            y index=3, 
            col sep=comma] {csv/fresh/p.csv};

        \addplot
        table [
            x index=0, 
            y index=2, 
            col sep=comma] {csv/fresh/p.csv};

        \addplot
        table [
            x index=0, 
            y index=1, 
            col sep=comma] {csv/fresh/p.csv};

        \addlegendentryexpanded{$p= 100\%$};
        \addlegendentryexpanded{$p= 10\%$};
        \addlegendentryexpanded{$p= 3\%$};
        \addlegendentryexpanded{$p= 1\%$};
        \addlegendentryexpanded{$p= 0.3\%$};

    \nextgroupplot[
        title = {$p = 10\%$},
        xmax=10000,
        xmode = log,
        ymode = log,
        ylabel={WD},
        xlabel={$n$},
        axis x line*=bottom,
        axis y line*=left,
        grid,
        legend style={
                at={(0.4,.3)},
                anchor=east,
                font=\footnotesize},
                legend style={nodes={scale=\legendscale, transform shape}}
        ]
        
        \addplot
        table [
            x index=0, 
            y index=4, 
            col sep=comma] {csv/fresh/p.csv};

        \addplot
        table [
            x index=0, 
            y index=4, 
            col sep=comma] {csv/fresh/r.csv};

        \addplot
        table [
            x index=0, 
            y index=3, 
            col sep=comma] {csv/fresh/r.csv};

        \addplot
        table [
            x index=0, 
            y index=2, 
            col sep=comma] {csv/fresh/r.csv};

        \addplot
        table [
            x index=0, 
            y index=1, 
            col sep=comma] {csv/fresh/r.csv};

        \addlegendentryexpanded{$\bias = 1$};
        \addlegendentryexpanded{$\bias = 0.97 $};
        \addlegendentryexpanded{$\bias = 0.95$};
        \addlegendentryexpanded{$\bias = 0.93$};
        \addlegendentryexpanded{$\bias = 0.9$};

\end{groupplot}

\end{tikzpicture}

\caption{ The Wasserstein distance (WD) as the whole dataset size increases for different values of $p$ (left), and sampling bias (right). }
\label{fig:fresh_appendix_1}

\end{figure}

\textbf{Experiment 2}: Here we assume that we are sampling from an environment where $p$ percent of data is real, and the rest is synthetic data from the previous generation $\model^{t-1}$ with sampling bias $\bias$. We change the total number of data in the dataset $n = |\mathcal{D}^t|$, with $n_r = p \times n$ and $n_s = (1-n) \times p$. We show the Wasserstein distance for different $p$ and $\bias$ in Figure \ref{fig:fresh_appendix_1}. 

Let us first examine the dynamics of the Gaussian \fresh{} without sampling bias ($\bias=1$). We observe in \cref{fig:fresh_appendix_1} (left) that the Wasserstein distance (WD) decreases with respect to dataset size $n$. However, the presence of synthetic data ($p < 100\%$) decreases the rate at which the WD decreases, and increases the overall WD each generation in the \fresh{}. \textit{This means that with presence of synthetic data in the Internet, the progress of generative models will become slower}

In the presence of sampling bias ($\bias<1$, \cref{fig:fresh_appendix_1} right), we see that even for close values of $\bias$ to 1, the Wasserstein distance follows a sub-linear trend, meaning that eventually the rate of progress in generative models will effectively stop, no matter how much (realistically) the total dataset size is increased.

\clearpage

\bibliography{bibliography}

\end{document}